%% file: main.tex
\pdfoutput=1
\documentclass{article}

\usepackage{iclr2027_preprint,times}

\input{preamble}
\input{macros}

\newcommand{\Vbar}{\bar{V}}
\newcommand{\Xbar}{\bar{X}}
\newcommand{\xbar}{\bar{x}}
\newcommand{\cAx}{\cA_{\sf X}}
\newcommand{\cAv}{\cA_{\sf V}}
\newcommand{\Fx}{F_{\sf X}}
\newcommand{\Fv}{F_{\sf V}}
\newcommand{\Fbar}{\bar{F}}
\newcommand{\Fbarx}{\Fbar_{\sf X}}
\newcommand{\Fbarv}{\Fbar_{\sf V}}

\title{Learning to Restart: Differentiable Restart Policies for Optimization Dynamics}

\author{
Yiman Fong$^{1}$ \qquad
Zhonglin Xie$^{2}$ \qquad
Zaiwen Wen$^{2}$ \\[0.8em]
\small $^{1}$Department of Industrial Engineering, Tsinghua University \\
\small $^{2}$Beijing International Center for Mathematical Research, Peking University \\[0.5em]
\small\ttfamily fangym23@mails.tsinghua.edu.cn \quad
zlxie@pku.edu.cn \quad
wenzw@pku.edu.cn
}

\begin{document}

\maketitle

\begin{abstract}
Adaptive restart mechanisms, which reset the momentum or other auxiliary state of an iterative optimizer, play a central role in the empirical performance of accelerated first-order methods, yet existing restart rules remain almost entirely hand-designed. The obstruction to learning restart policies is that, unlike the typical Learning to Optimize (L2O) setting, the discrete restart events make the resulting trajectory non-differentiable in the algorithm parameters and thus block gradient-based optimization of the restart policy.
We resolve this obstruction by developing a differentiable framework for restart policies through their continuous-time limit. We introduce \emph{surrogate gradients} for the restart events and optimization trajectories that are efficient to compute and provably converge to their continuous-time counterparts. We validate the effectiveness of our surrogate gradients in L2O settings, training restart policies that improve upon standard handcrafted adaptive restart criteria on quadratic and linear programs, and that transfer zero-shot from synthetic quadratics to logistic regression on real datasets.
\end{abstract}

\section{Introduction}
\label{sec:intro}

Optimization algorithms underlie a wide range of modern applications, including operations research~\citep{taha1997operations}, the training of large-scale machine learning models~\citep{zhao2025surveylargelanguagemodels}, and decision-making in finance~\citep{pilbeam2018finance}. Substantial effort has therefore been devoted to accelerating these algorithms, both by analytical design and, more recently, by data-driven approaches such as hyperparameter optimization~\citep{feurer2019hyperparameter} and learning to optimize (L2O)~\citep{chen2022learning,chen2024learning}. In this line of work, the training objective is differentiable in the parameters to be optimized, which enables efficient gradient-based optimization.

What is largely missing from this data-driven toolbox is the \emph{restart} mechanism: adaptive resets of the momentum or other auxiliary state of an iterative algorithm. Restart is routinely used to accelerate convergence on a wide range of convex optimization problems~\citep[etc.]{o2015adaptive,giselsson2014monotonicity,fercoq2019adaptive}, and it is also a key ingredient behind the empirical performance of PDLP on large-scale linear programs~\citep{applegate2021practical,applegate2023faster,lu2025geometry,applegate2025pdlp}. The vast majority of existing restart rules, however, are designed analytically in a case-by-case manner. It is therefore natural to follow the L2O paradigm and \emph{learn} the restart criterion, by parameterizing it via a neural network and optimizing it directly against a downstream optimization objective. The key obstacle is that the restart events are discrete in time, so the resulting optimization trajectory is \emph{non-differentiable}, which precludes a naive application of gradient-based learning.

In this work, we resolve this obstacle by developing a differentiable framework for learning restart rules in optimization dynamics. Our starting point is the observation that, although the restart events of a discrete-time algorithm are non-differentiable in the algorithm parameters, the restart events in its continuous-time limit dynamics \emph{are} differentiable. Our main contributions are summarized as follows:
\begin{itemize}[leftmargin=1.5em,itemsep=0.15em]
    \item \textbf{Continuous-time differentiable restart.} In continuous-time dynamics, we apply the implicit function theorem to obtain closed-form representations of the gradients for both the restart times and the trajectory (\cref{thm:restart_time_sensitivity}). We demonstrate that by using these gradient surrogates,  a parameterized restart rule can be optimized to outperform the standard hand-designed adaptive restart criteria on the underlying ODE dynamics (\cref{sec:qp-validation}).
    \item \textbf{Efficient discrete approximation.} We develop a \emph{surrogate gradient} (\cref{def:disc_restart_sensitivity})---the discrete-time analogue of the continuous restart sensitivity that can be implemented efficiently (e.g., through the standard automatic-differentiation framework). We prove that, whenever the discrete dynamics are \emph{consistent} with their continuous-time limit, the surrogate gradients converge to the continuous gradients (\cref{thm:disc_cont_close}).
    \item \textbf{Applications to L2O.} We instantiate the framework in the L2O setting through experiments on Quadratic Programming (\cref{sec:qp-discrete}) and on Linear Programming (\cref{sec:pdlp}), each with a synthetic training family and real test problems. The learned restart policies consistently outperform the standard handcrafted restart criteria on both problem families, confirming the effectiveness of our discrete surrogate gradients. In particular, a policy trained only on synthetic quadratics transfers zero-shot to logistic regression on real datasets and to tomographic reconstruction from real scanner measurements (\cref{sec:logreg}), where it improves over the best handcrafted criterion.
\end{itemize}

\input{related_work.tex}

\section{Differentiable Restarts: A Continuous-time Perspective}
\label{sec:problem_setup}

In the following, we first motivate the technical obstacle of applying the L2O paradigm to learn the adaptive restart policies.

\paragraph{A motivating example}
Consider NAG with parameter-dependent stepsize $\alpha_k(\theta)$ and momentum coefficient $\beta_k(\theta)$ applied to a smooth convex objective $f$:
\begin{align}\label{eq:NAG-parametric}
    \textstyle
    x_{(k+1)} = y_{(k)} - \alpha_k(\theta)\,\nabla f(y_{(k)}),
    \qquad
    y_{(k+1)} = x_{(k+1)} + \beta_k(\theta)\bigl(x_{(k+1)} - x_{(k)}\bigr),
\end{align}
where both parameter schedules are differentiable in the hyperparameter $\theta$. Without restarts, the iterate $x_{(N)}$ produced after $N$ steps is a smooth function of $\theta$, and any objective built from it---for instance $f(x_{(N)})$---admits ordinary gradient-based optimization in $\theta$.

To equip~\eqref{eq:NAG-parametric} with an adaptive restart in the spirit of \citet{o2015adaptive}, the natural construction is to introduce a parameterized restart score $\Vbar(x,y;\theta_{\rm rst})$ and trigger a restart whenever $\Vbar(x_{(k)},y_{(k)};\theta_{\rm rst})\leq 0$, at which point the momentum is reset by setting $y_{(k)}\leftarrow x_{(k)}$. The full parameter vector is then $(\theta,\theta_{\rm rst})$, and the L2O objective becomes to learn the within-phase schedule and the restart criterion jointly.
However, the indices at which $\Vbar$ crosses below zero are \emph{integer-valued} and locally constant in $(\theta,\theta_{\rm rst})$ away from decision boundaries. The issue is particularly severe for $\theta_{\rm rst}$, which influences the trajectory \emph{only} through the location of the restart events: away from decision boundaries, $x_{(N)}$ is locally constant in $\theta_{\rm rst}$, so its derivative vanishes and carries no information.\footnote{The straight-through estimator~\citep{bengio2013estimating} corresponds to setting the integer-step derivative to zero and would therefore produce $\partial_{\theta_{\rm rst}} x_{(N)}\equiv 0$ identically in this setting, providing no gradient information for $\theta_{\rm rst}$.}

\paragraph{Our approach}
To circumvent this obstruction, we adopt the continuous-time perspective of optimization algorithms. Most discrete optimizers (including NAG and the primal--dual schemes considered later) admit a continuous-time limit (e.g., a gradient flow) in which the restart triggered by $\Vbar = 0$ becomes \emph{smooth}. In this section, we formalize this intuition, derive the continuous-time gradients, and validate their effectiveness. \Cref{sec:discrete} then transfers these continuous-time gradients back to the discrete algorithm via an efficient surrogate, closing the loop between the continuous-time analysis and the practical iterative optimizers.

\subsection{ODE Dynamics and Restarts}

We use the word \emph{state} broadly: depending on the algorithm, it may comprise the optimization variable, momentum variables, primal--dual pairs, running averages, or any other auxiliary quantity maintained by the optimizer.
The evolution of the state in the dynamical systems of interest can typically be written as a parameterized ordinary differential equation (ODE):
\begin{equation}\label{eq:ODE}
    \dot{X}(t)=-\cAx(X(t),t;\theta), \qquad X(0)=X_0,
\end{equation}
where $\cAx:\RR^D\times\Theta\to\RR^D$ encodes the instantaneous direction of state motion induced by the algorithm with (hyper-)parameter $\theta\in\Theta$.

Many classical optimization algorithms (e.g., gradient descent, momentum methods, accelerated gradient methods, and primal--dual gradients) admit such a gradient-flow representation, in which the state $X$ collects the iterate together with any momentum or auxiliary variables maintained by the algorithm and the field $\cAx$ encodes the corresponding update rule. We illustrate the abstraction with three concrete examples below.

\paragraph{Gradient flow.}
The simplest example is the (rescaled) gradient flow,
\[
\textstyle
\dot{x}(t)=-\alpha(t;\theta)\,\nabla f(x(t)),
\]
where $f$ is the objective and the scaling function $\alpha(t;\theta)$ is time- and parameter-dependent. In the form of \eqref{eq:ODE}, the field is $\cAx(x,t;\theta)=\alpha(t;\theta)\,\nabla f(x)$.

\paragraph{Nesterov's accelerated gradient (NAG).}
Several ODE models of NAG\footnote{The corresponding discrete iteration and the standard hand-crafted restart rules are summarized in \cref{app:algorithms_with_restarts}.} have been proposed in the literature~\citep{su2016differential,shi2022understanding}. On the smooth convex minimization problem $\min_{x\in\RR^d} f(x)$, the limiting behaviour of NAG with stepsize $s>0$ can be captured by the \emph{high-resolution} ODE~\citep{shi2022understanding}:
\begin{align}\label{eq:NAG-C-ODE}
    \textstyle
\ddot{x}(t)+\frac{3}{t+\tau}\dot{x}(t)
+\sqrt{s}\,\nabla^2 f(x(t))\,\dot{x}(t)
+\left(1+\frac{3\sqrt{s}}{2(t+\tau)}\right)\nabla f(x(t))=0,
\end{align}
with $\tau=3\sqrt{s}/2$, $x(0)=x_0$ and $\dot{x}(0)=-\sqrt{s}\nabla f(x_0)$,
which is shown to accurately track the trajectory of the discrete iteration. This second-order ODE fits the abstract form~\eqref{eq:ODE} if the state $X(t)=[x(t);v(t)]$ consists of the iterate and the momentum $v(t)=\dot{x}(t)$.
We use these dynamics throughout \cref{sec:qp-validation} to validate the continuous-time gradients derived in \cref{sec:diff_restart}.

\paragraph{Primal--dual gradient dynamics.}
For minimax (primal--dual) problems $\min_x \max_y \cL(x,y)$, the analogous continuous-time model is the primal--dual gradient flow~\citep{arrow1958studies,feijer2010stability},
\[
\textstyle
    \dot{x}(t)=-\nabla_x \cL(x(t),y(t)), \qquad
    \dot{y}(t)=\nabla_y \cL(x(t),y(t)).
\]
Closely related dynamics have been studied for constrained minimax problems~\citep{feijer2010stability,cherukuri2016asymptotic} and for accelerated primal--dual methods such as the primal--dual hybrid gradient (PDHG) method~\citep{chambolle2011first,li2024understanding}.

\subsection{Dynamics with Restarts}
\label{sec:dynamics_with_restart}

ODE dynamics with restarts can then be (informally) described by augmenting the ODE~\eqref{eq:ODE} with an additional scalar score $V(t)$ that determines the restarting (with $x(0)=x_0$ and $V(0)=1$):
\begin{equation}\label{eq:ODE-restart}
\begin{cases}
    \frac{dX(t)}{dt}=-\cAx(X(t),t;\theta), \\[2pt]
    \frac{dV(t)}{dt}=-\cAv(X(t),V(t),t;\theta),
\end{cases}\qquad \text{while }V(t)>0.
\end{equation}
At any time $t$ such that $V(t-)=0$, a restart is triggered: we reset $V(t)=1$ and apply a restart map $R:\RR^D\to\RR^D$ to obtain $X(t)=R(X(t-))$. Here $\cA=(\cAx,\cAv)$ specifies the joint dynamics of the state and the restart score, while $R$ specifies the restart operation, which may, for instance, leave the iterate component unchanged but reset the momentum variables to zero.

The system in~\eqref{eq:ODE-restart} is hybrid: between consecutive restarts the state evolves continuously according to a smooth ODE, while at each restart it is reinitialized in a discontinuous fashion via the map $R$ together with the score reset $V\!\leftarrow\!1$. Such hybrid evolutions can be analyzed rigorously within the framework of hybrid dynamical systems~\citep{van2007introduction,goebel2009hybrid} or neural event functions for ODEs~\citep{chen2020learning}. In the following, we develop the minimal notation required to describe such systems.

\paragraph{General flow-map dynamics.}
To describe dynamics with restarts, we start
by specifying the evolution within each restart \emph{phase}. Let
\[
    F=(\Fx,\Fv):\RR_+\times\RR^D\times\Theta\to\RR^{D}\times \RR
\]
be a differentiable flow map, with $\Fx(0;x,\theta)=x$ and $\Fv(0;x,\theta)>0$ at each phase initialization, where $\Fx(t;x,\theta)$ ($\Fv(t;x,\theta)$) denotes the state $X$ (score $V$) reached after time $t$ when the phase starts from $x$. In the ODE setting above, $F(t;x,\theta)$ is the solution $(X(t),V(t))$ at time $t$ of ODE~\eqref{eq:ODE-restart} initialized at $X(0)=x$ and $V(0)=1$, but our framework does not require $F$ to be of this specific form.

Intuitively, the whole trajectory is obtained by repeatedly flowing under $F$ until the learned event condition $V\le 0$ is met, and then starting the next phase from the resulting state.
Formally, we consider a parameterized global trajectory $(X(t)=X(t;\theta))_{t\geq 0}$ obtained by concatenating phases. On the $i$-th phase $[T_i,T_{i+1})$, the trajectory follows
\begin{equation}
    X(T_i+t)=\Fx(t;X(T_i),\theta), \qquad 0\leq t<L_i:= T_{i+1}-T_i,
    \label{eq:local_flow}
\end{equation}
where $L_i$ denotes the length of the phase determined by the score $V(T_i+t):=\Fv(t;X(T_i),\theta)$ via
\begin{equation}
    L_i:= \inf\{t\geq 0: \Fv(t;X(T_i),\theta)\le 0\},\qquad \inf\varnothing:=+\infty.
    \label{eq:phase_length}
\end{equation}
At a finite event, the pre-reset score satisfies $V(T_{i+1}-)=\Fv(L_i;X(T_i),\theta)=0$, and the post-reset state is $X(T_{i+1})=R(\Fx(L_i;X(T_i),\theta))$. If $L_i=+\infty$, the current phase continues without a further reset.

We summarize this construction in the following definition.

\begin{definition}[Continuous restarted trajectory]
\label{def:continuous_restarted_trajectory}
The restarted trajectory $(X(t))_{t\geq 0}$ is defined recursively, starting with $T_0=0$ and initial point $X(0)=x_0$. For each phase $i$ with starting time $T_i$ and initial point $x_i:= X(T_i)$, we define
\begin{equation}
    X(T_i+t)=\Fx(t;x_i,\theta), \qquad 0\leq t< L_i,
    \label{eq:restart_piecewise_definition}
\end{equation}
with restart time $L_i$ defined in \eqref{eq:phase_length}. For finite $L_i$, we then set
\begin{equation}
    T_{i+1}=T_i+L_i, \qquad
    x_{i+1}=X(T_{i+1})=R(\Fx(L_i;x_i,\theta)).
    \label{eq:restart_recursion}
\end{equation}
\end{definition}

The notation here is deliberately compact, but it straightforwardly covers the standard restart rules used in practice. As a concrete instance, suppose that the restart score depends only on the elapsed time $t$, the current state $X(T_i+t)$, and the state $x_i=X(T_i)$ at the start of the current phase. Then $\Fv$ admits the form
\begin{equation}
    \textstyle
    \Fv(t;x_i,\theta)
    =\bar V\bigl(t,X(T_i+t),x_i;\theta\bigr),
    \label{eq:V_bar_notation}
\end{equation}
where $\bar V$ is a parameterized function of the elapsed time, the current state, and the phase-start state. Handcrafted restart rules---such as the function-value and gradient-based rules of \citet{o2015adaptive}---correspond to such fixed, problem-specific choices of $\bar V$. In our setting, $\bar V$ is parameterized by a small neural network and its parameters are learned jointly with any parameters of the within-phase dynamics.

\subsection{Differentiating Restart}
\label{sec:diff_restart}

The restart time is defined implicitly by the event equation $\Fv(L_i;x_i,\theta)=0$, and its derivative can therefore be derived through the chain rule and the implicit function theorem, as in the differentiation of stopping times~\citep{chen2020learning,xie2025accelerating}. \footnote{
    Throughout, $\partial_\theta$ applied to a phase quantity such as $L_i$, $x_i$, or $X(t)$ denotes the \emph{total} derivative with respect to $\theta$, while $\partial_\theta F$, $\partial_\theta V$ denote the partial derivatives of the flow and the score in their explicit $\theta$ argument; $\partial_x$ and $\partial_t$ are interpreted similarly.
}

\newcommand{\Fr}{F_{\sf R}}
\newcommand{\Fbarr}{\Fbar_{\sf R}}
\begin{theorem}[Sensitivity of learnable restart dynamics]
\label{thm:restart_time_sensitivity}
Fix a finite horizon $T$ and an initial state $x_0$. Suppose $F$ and $R$ are $C^1$. At $\theta_0$, assume there are exactly $q<\infty$ restarts in $[0,T]$, occurring at $0<T_1<\cdots<T_q<T$. For every realized event $0\leq i<q$, assume $\partial_t\Fv(L_i;x_i,\theta_0)<0$. Then $L_i$ and $x_{i+1}$ are $C^1$ in a neighborhood of $\theta_0$, with recursion
\begin{equation}
    \partial_\theta L_i
    =
    -
    \frac{
        \partial_\theta \Fv(L_i;x_i,\theta)
        +
        \partial_x \Fv(L_i;x_i,\theta)\cdot \partial_\theta x_i
    }{
        \partial_t \Fv(L_i;x_i,\theta)
    },
    \label{eq:dLi_dp}
\end{equation}
and
\begin{equation}
    \partial_\theta x_{i+1}
    =
    \partial_t \Fr(L_i;x_i,\theta)\,\partial_\theta L_i
    +
    \partial_x \Fr(L_i;x_i,\theta)\,\partial_\theta x_i
    +
    \partial_\theta \Fr(L_i;x_i,\theta),
    \label{eq:dxi1_dp}
\end{equation}
where $\partial_x \Fv$ denotes the derivative of the score-flow component $\Fv$ with respect to the phase initialization $x$ (i.e., the second argument), and we denote $\Fr:= R\circ \Fx$.
\end{theorem}

Consequently, for a fixed global time $t\in[0,T]$ with $t\in(T_i,T_{i+1})$, the state $X(t)=\Fx(t-T_i;x_i,\theta)$ is also differentiable with respect to $\theta$, with gradient
\begin{equation}
    \partial_\theta X(t)
    =
    -\partial_t \Fx(t-T_i;x_i,\theta)\,\partial_\theta T_i
    +\partial_x \Fx(t-T_i;x_i,\theta)\,\partial_\theta x_i
    +\partial_\theta \Fx(t-T_i;x_i,\theta),
    \label{eq:dXt_dp}
\end{equation}
where $\partial_\theta T_i=\sum_{j=0}^{i-1} \partial_\theta L_j$ by definition. The first term is a boundary-shift correction: at a fixed global time, perturbing $\theta$ moves the earlier restart boundaries, which changes the local time corresponding to $t$. A full proof of \cref{thm:restart_time_sensitivity} based on a phase-by-phase induction with the implicit function theorem is deferred to \cref{app:restart_sensitivity_derivation}.

In the remainder of this section, we validate the effectiveness of the gradients.
In \cref{sec:discrete}, we then develop a discrete-time counterpart of these continuous-time sensitivities that retains the same structure while avoiding the cost of exact ODE differentiation.

\subsection{Validation: Optimizing the Restarts in the Continuous-time Dynamics}
\label{sec:qp-validation}

\newcommand{\MLP}{\mathsf{MLP}}

In the following, we demonstrate the effectiveness of the restart gradients of \cref{thm:restart_time_sensitivity} by optimizing the restarts for the ODE~\eqref{eq:NAG-C-ODE}. We consider QP problems of the form
\begin{equation}\label{eq:QP}
\textstyle
\min_{x\in\RR^d} f(x)=\tfrac12 x^\top A x-b^\top x+c
\end{equation}
with dimension $d=30$ and condition number $\kappa=80$; the precise generation protocol of $f$ is deferred to \cref{app:QP}. On such problems the ODE~\eqref{eq:NAG-C-ODE} and the corresponding sensitivity equations can be simulated with high accuracy.

We parameterize the learnable restart score via a small MLP $V(t;\theta)=\MLP(\phi(x(t));\theta)$, where $\phi$ is a simple feature map (see \cref{app:QP-ODE}). The parameter $\theta$ is optimized with respect to the following loss (integrated function value):
\begin{equation}\label{eq:qp-loss}
    J(\theta)\;:=\;\int_0^T f(x(t;\theta))\,dt,
\end{equation}
with gradients computed through~\eqref{eq:dLi_dp}--\eqref{eq:dXt_dp}.
We compare the learned policy against two handcrafted continuous-time restart criteria, which are continuous analogues of the standard discrete adaptive-restart rules of~\citet{o2015adaptive}:
\begin{itemize}[leftmargin=1.5em,itemsep=0.15em]
    \item \textbf{Function restart}: $V_{\sf func}(t)=-\nabla f(x(t))^\top \dot x(t)$, the continuous analogue of the function-based rule ``restart iff $f(x_{k+1})>f(x_k)$''.
    \item \textbf{Gradient restart}: $V_{\sf grad}(t)=-\nabla f(x(t)+\sqrt{s}\,\dot x(t))^\top \dot x(t)$, the continuous analogue of the gradient-based rule ``restart iff $\nabla f(y_k)^\top(x_{k+1}-x_k)>0$''.
\end{itemize}

\begin{figure}[t]
\centering
\includegraphics[width=0.25\linewidth]{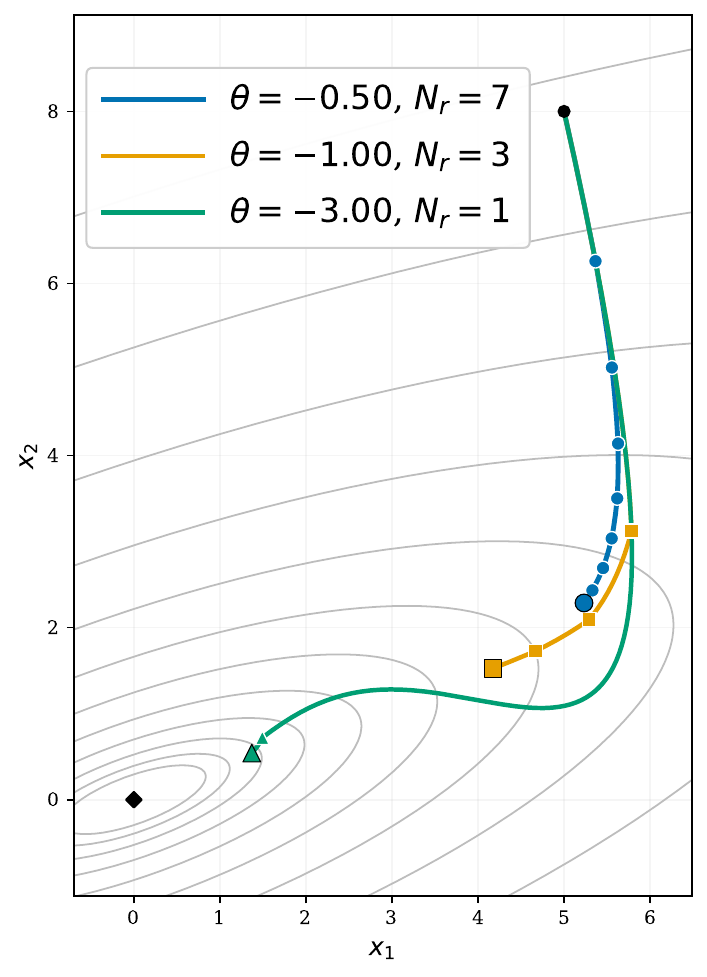}~~~~~~~~~~
\includegraphics[width=0.52\linewidth]{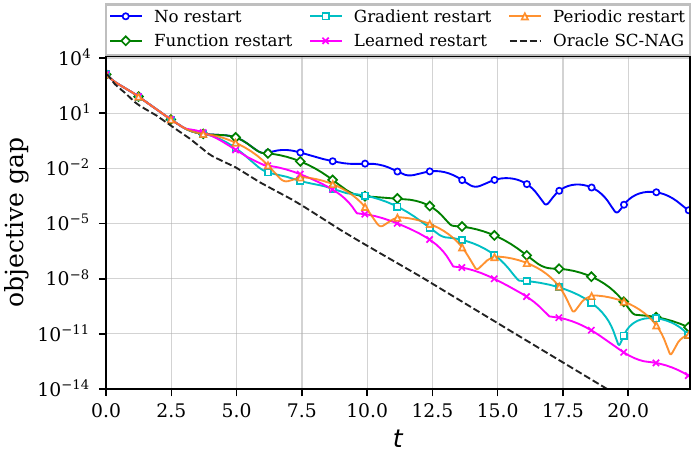}
\caption{Continuous-time validation on the QP family of \eqref{eq:QP}. \emph{Left:} ODE trajectories induced by \eqref{eq:NAG-C-ODE} under different restart policies on a 2-dimensional quadratic instance, illustrating how the learned policy reshapes the iterate path. \emph{Right:} objective gap $f(x(t))-\min f$ along the optimization trajectories induced by the hand-designed and learned restart policies ($d=30$, $\kappa=80$). The learned restart policy consistently improves over both hand-designed criteria and approaches the strongly-convex reference.
}
\label{fig:qp-validation}
\end{figure}

\Cref{fig:qp-validation} (\emph{right}) reports the objective gap along the optimization trajectories induced by the hand-designed restart policies and the learned restart policy. The learned restart policy achieves a better final gap $f(x(t))-\min f$ than the baseline restart criteria, and its convergence is more stable. As a reference-only ceiling, we also plot the strongly-convex accelerated ODE that uses the true strong-convexity parameter $\mu$ (detailed in \cref{app:QP-ODE}).\footnote{This last curve relies on knowing $\mu$; adaptive restart is precisely the mechanism that recovers the same exponential convergence rate without such knowledge.} This validates the learnability of the restart criterion in the continuous-time regime via the gradients derived in \cref{sec:diff_restart}; more extensive experiments in the discrete-time setting are deferred to \cref{sec:qp-discrete}.

\section{Efficient Discrete Approximation of Restart Sensitivities}\label{sec:discrete}

\newcommand{\grad}{\mathsf{grad}}
\newcommand{\sgrad}{\mathsf{stopgrad}}

\newcommand{\disc}[1]{\grad(#1)}

To apply the insights of \cref{sec:diff_restart} to practical optimization algorithms, which operate in \emph{discrete} time, we now develop the discrete analogue of the continuous-time restart gradients. Concretely, in this section we introduce a notion of \emph{surrogate gradients} for both the restart step (\cref{def:disc_restart_sensitivity}) and the restarted trajectory in the discrete setting, and we show that the surrogate gradients closely approximate the continuous-time gradient of \cref{def:continuous_restarted_trajectory} whenever the discrete dynamics are sufficiently close to the continuous-time dynamics.

Following the notation in \cref{sec:problem_setup}, we assume that the discrete-time dynamics are described by a flow map $\Fbar=(\Fbarx,\Fbarv)$ as follows.

\begin{definition}[Discrete restarted trajectory]
\label{def:discrete_restarted_trajectory}
The restarted discrete trajectory $(\Xbar(n)=\Xbar(n;\theta))_{n\geq 0}$ is defined recursively, starting with $N_0=0$ and $\xbar_0=x_0$. Assume $\Fbarx(0;x,\theta)=x$ and $\Fbarv(0;x,\theta)>0$ at every phase initialization. For each phase $i$ with starting step $N_i$ and initial point $\xbar_i:=\Xbar(N_i)$, define
\begin{equation}
    \Xbar(N_i+k)=\Fbarx(k;\xbar_i,\theta), \qquad \Vbar(N_i+k)=\Fbarv(k;\xbar_i,\theta),
    \label{eq:disc_restart_piecewise_definition}
\end{equation}
for each $0\leq k<S_i$, where the discrete restart time $S_i$ is defined as
\begin{equation}
    S_i:=\min\crl*{k\geq 1:\Fbarv(k;\xbar_i,\theta)\leq 0},\qquad \min\varnothing:=+\infty.
\end{equation}
For finite $S_i$, we then set $N_{i+1}=N_i+S_i$ and $\xbar_{i+1}=\Xbar(N_{i+1})=R(\Fbarx(S_i;\xbar_i,\theta))$.
\end{definition}

As we have mentioned, the integer-valued restart step $S_i$ is locally constant in $\theta$, and therefore does not admit a non-trivial ordinary derivative. To define a meaningful surrogate, we assume that $\Fbar$ is a discretization of a continuous-time system in the sense of \cref{def:continuous_restarted_trajectory}: we consider the time grid $t_n=nh$ for $n=0,1,\dots,N$, where $h>0$ is the discretization stepsize and $N$ is the total number of steps with $T=Nh$.
Then, following \cref{def:continuous_restarted_trajectory}, we can introduce the following surrogate gradients.

\begin{definition}[Surrogate gradient]
\label{def:disc_restart_sensitivity}
For a finite first crossing $S_i$, let $\Delta_i:=\Fbarv(S_i;\xbar_i,\theta)-\Fbarv(S_i-1;\xbar_i,\theta)$, which satisfies $\Delta_i<0$. Given $\disc{\xbar_i}$, define the formal step sensitivity $\disc{S_i}$ by
\begin{equation}
    \disc{S_i}
    =
    -
    \frac{
        \partial_\theta \Fbarv(S_i;\xbar_i,\theta)
        +
        \partial_x \Fbarv(S_i;\xbar_i,\theta)\disc{\xbar_i}
    }{
        \Delta_i
    }.
\label{eq:disc_dSi_dp}
\end{equation}
The restart-state sensitivity is defined by
\begin{equation}
 \begin{aligned}
    \disc{\xbar_{i+1}}
    =&~
    \bigl(\Fbarr(S_i;\xbar_i,\theta)-\Fbarr(S_i-1;\xbar_i,\theta)\bigr)\disc{S_i} \\
   &~+
    \partial_x \Fbarr(S_i;\xbar_i,\theta)\disc{\xbar_i}
    +
    \partial_\theta \Fbarr(S_i;\xbar_i,\theta),
 \end{aligned}
    \label{eq:disc_dxi1_dp}
\end{equation}
where we denote $\Fbarr:=R\circ \Fbarx$.
For interior global steps $N_i<n<N_{i+1}$, the global-trajectory sensitivity is defined by the discrete analogue of \eqref{eq:dXt_dp},
\begin{align}
        \disc{\Xbar(n)}
    =&~
    -\bigl(\Fbarx(n-N_i;\xbar_i,\theta)-\Fbarx(n-1-N_i;\xbar_i,\theta)\bigr)\disc{N_i}\label{eq:disc_dXn} \\
    &~ +
    \partial_x \Fbarx(n-N_i;\xbar_i,\theta)\disc{\xbar_i}
    +
    \partial_\theta \Fbarx(n-N_i;\xbar_i,\theta),\notag
\end{align}
where $\disc{N_i}=\sum_{j<i}\disc{S_j}$ accumulates the boundary-shift surrogate gradients from previous phases.
\end{definition}

\paragraph{Intuition} We can mimic the continuous derivation in \cref{sec:diff_restart} and treat the discrete event equation $\Fbarv(S_i;\xbar_i,\theta)\approx 0$ as an implicit constraint and apply the implicit function theorem formally, replacing the derivative with respect to the (integer) step index by a backward difference. Symbolically, using the chain rule,
\begin{align*}
    0
    \;\approx\;&~
    \nabla_\theta \Fbarv(S_i;\xbar_i,\theta)
    \;\approx\;
    \partial_k \Fbarv(S_i;\xbar_i,\theta)\cdot \partial_\theta S_i
    +
    \partial_x \Fbarv(S_i;\xbar_i,\theta)\cdot \partial_\theta \xbar_i
    +
    \partial_\theta \Fbarv(S_i;\xbar_i,\theta) \\
    \approx &~ \prn{\Fbarv(S_i;\xbar_i,\theta)-\Fbarv(S_i-1;\xbar_i,\theta)}\cdot \partial_\theta S_i
    +
    \partial_x \Fbarv(S_i;\xbar_i,\theta)\cdot \partial_\theta \xbar_i
    +
    \partial_\theta \Fbarv(S_i;\xbar_i,\theta),
\end{align*}
where $\partial_k \Fbarv$, $\partial_\theta S_i$, and $\partial_\theta \xbar_i$ are formal symbols: they do not exist in the strict sense, since $S_i$ is integer-valued and $\xbar_i$ inherits this non-differentiability. Although this manipulation is purely formal, it exposes the right structure: $S_i$ plays the role of the continuous restart time $L_i$, with the time derivative $\partial_t V$ replaced by the discrete backward difference $\Delta_i$ defined below. Solving this formal equation for $\partial_\theta S_i$ then motivates \cref{def:disc_restart_sensitivity}. We emphasize that this derivation is purely intuitive and does not by itself justify the surrogate; the actual theoretical guarantee comes from connecting \cref{def:disc_restart_sensitivity} to the continuous-time gradients of \cref{def:continuous_restarted_trajectory}, which is the subject of \cref{thm:disc_cont_close}.

\paragraph{Computational efficiency}
The sensitivities defined in~\eqref{eq:disc_dSi_dp}--\eqref{eq:disc_dxi1_dp} are designed to be plugged directly into a standard automatic-differentiation framework (\cref{app:auto-diff}). \Cref{app:forward-implementation} gives a single-pass forward-sensitivity implementation, including the restart and boundary-shift updates, with storage independent of trajectory length.

\paragraph{Theoretical guarantees}
We next show that the surrogate gradients are \emph{accurate}: provided that the discrete flow map $\Fbar$ approximates $F$ in the quantitative sense of the following assumption, the surrogate gradients of the restart time, the restart state, and the global trajectory all converge to their continuous counterparts.

\begin{assumption}
\label{ass:disc_consistency}
Fix a finite horizon $T$ and a local-time interval $[0,H]$ with $H>T$. On a neighborhood of the realized phase initializations and parameters, assume that there are constants $\gamma,M>0$ such that the following holds.

(1) The event hypotheses of \cref{thm:restart_time_sensitivity} hold, with $q$ restarts in $[0,T]$, and $\partial_t \Fv(L_i;x_i,\theta)\leq -\gamma$ for $0\leq i<q$.

(2) Smoothness: $\nrm{\partial_{(t,x,\theta)} F}+\nrm{\nabla R}\leq M$, and $\partial_{(t,x,\theta)} F$ and $\nabla R$ are both $M$-Lipschitz.

(3) Consistency: for $t_k=kh$, the flow maps satisfy
\begin{align}
    \nrm{\Fbar(k;x,\theta)-F(t_k;x,\theta)}
    +\nrm{\partial_{(x,\theta)}\Fbar(k;x,\theta)-\partial_{(x,\theta)}F(t_k;x,\theta)}
    &\leq \epsilon_h, \label{eq:Fhat_consistency}\\
    \nrm*{
    \tfrac{1}{h}\prn*{\Fbar(k+1;x,\theta)-\Fbar(k;x,\theta)}
    -\partial_t F(t_k;x,\theta)}
    &\leq \epsilon_h, \label{eq:Vhat_time_consistency}
\end{align}
where $\epsilon_h\to 0$ as $h\to 0$.
\end{assumption}

\begin{theorem}[Closeness of surrogate gradients to continuous gradients]
\label{thm:disc_cont_close}
Suppose that Assumption~\ref{ass:disc_consistency} holds, $\xbar_0=x_0$ is independent of $\theta$, and $h$ is sufficiently small.
The discrete and continuous trajectories have the same $q$ restarts in $[0,T]$. For $0\leq i<q$, the surrogate gradient in \cref{def:disc_restart_sensitivity} satisfies
    \begin{align}
    \abs{hS_i-L_i}+\nrm{h \disc{S_i}-\partial_\theta L_i}&=O(\epsilon_h+h), \label{eq:sensitivity_close_step}\\
    \nrm{\xbar_{i+1}-x_{i+1}}+\nrm{\disc{\xbar_{i+1}}-\partial_\theta x_{i+1}}&=O(\epsilon_h+h). \label{eq:sensitivity_close_state}
\end{align}
For a sufficiently large constant $C$, grid time $t_n=nh\in[0,T]$ with $\operatorname{dist}(t_n,\{0,T_1,\ldots,T_q\})\geq C(h+\epsilon_h)$, it holds that
\begin{align}
    \nrm{\Xbar(n)-X(t_n)}+\nrm{\disc{\Xbar(n)}- \partial_\theta X(t_n)}&=O(\epsilon_h+h).\label{eq:sensitivity_close_traj}
\end{align}
\end{theorem}

The proof and a more general formulation allowing nonuniform regularity near phase starts are given in \cref{app:disc_cont_close_proof}.
Finally, in \cref{sec:disc_validation}, we empirically validate the surrogate gradient of \cref{def:disc_restart_sensitivity} against its continuous-time counterpart in a setting where both can be computed accurately.

{
\paragraph{Policy improvement guarantees}
Built upon our framework, we establish second-order smoothness of the optimization dynamics with respect to the restart policy parameter (Appendix~\ref{thm:monotone-improvement}). In particular, projected gradient descent (PGD) on the policy parameter with an appropriate stepsize is guaranteed to improve, or at least preserve, performance on the problems used for training. Therefore, if the restart policy is initialized at a given restart method, the learned policy is guaranteed not to perform worse than the given method on the empirical fixed-horizon training objective.

In addition, the convergence guarantees of the restarted dynamics can be inherited from the optimization backbone. Therefore, the learned policy can exploit accelerated NAG/PDLP dynamics when beneficial, while the worst case is bounded by the convergence guarantee of vanilla gradient methods.

}

\input{experiment.tex}

\section{Conclusion}

We developed a differentiable framework for learning restart policies. Restart events are non-differentiable in discrete time but differentiable in the continuous-time limit; we derived the continuous-time gradients in closed form, introduced a discrete surrogate gradient that fits into standard automatic differentiation, and proved that it converges to the continuous-time gradient. Learned restart policies outperform handcrafted restart criteria on quadratic and linear programs, and transfer to unseen problem sizes and to real problems without retraining.

\bibliographystyle{iclr2027_conference}
\bibliography{ref}

\newpage
\appendix

\section{Additional Details}

\subsection{Nesterov's Accelerated Gradient with Restarts}
\label{app:algorithms_with_restarts}

On the smooth convex minimization problem $\min_{x\in\RR^d} f(x)$ with stepsize $s>0$, the standard NAG iteration with momentum coefficient $\beta_k=(k-1)/(k+2)$ can be written,\footnote{An equivalent formulation uses the auxiliary sequence $t_{k+1} = ({1+\sqrt{1+4t_k^2}})/{2}$ with $\beta_k = ({t_k-1})/{t_{k+1}}$, which yields the same asymptotic behaviour.} in terms of the iterate $x_k$ and a velocity-like momentum variable $v_k:= (x_k-x_{k-1})/\sqrt{s}$, as
\begin{equation}
\label{eq:app-nagc-iteration}
\begin{aligned}
    y_k       &= x_k+\beta_k\sqrt{s}\,v_k, \\
    x_{k+1}   &= y_k-s\,\nabla f(y_k), \\
    v_{k+1}   &= (x_{k+1}-x_k)/\sqrt{s}.
\end{aligned}
\end{equation}
The state in the abstract flow-map formulation is the pair $\xbar=(x,v)$, and the restart map zeroes the velocity:
\begin{equation}
    R(x,v)=(x,0).
    \label{eq:app-nagc-restart-map}
\end{equation}
With $h=\sqrt{s}$, the iteration~\eqref{eq:app-nagc-iteration} is a discretization of the high-resolution ODE~\eqref{eq:NAG-C-ODE} of \citet{shi2022understanding}; consistency for initial-position derivatives and time differences away from phase starts is verified in \cref{app:nag_consistency}.

The within-phase update~\eqref{eq:app-nagc-iteration} is augmented with a restart score $\Vbar(\,\cdot\,;\theta_{\rm rst})$ that triggers a reset whenever it crosses zero. Algorithm~\ref{alg:nagc-restart} summarizes the resulting iteration.

\begin{algorithm}[t]
\caption{NAG with adaptive restart.}
\label{alg:nagc-restart}
\begin{algorithmic}[1]
\REQUIRE Initial point $x_0\in\RR^d$, stepsize $s>0$, restart score $\Vbar(\,\cdot\,;\theta_{\rm rst})$, horizon $N$.
\STATE Set $x_{-1}\leftarrow x_0$, $v_0\leftarrow 0$, phase counter $k_{\rm rst}\leftarrow 0$, anchor $x_{\rm anc}\leftarrow x_0$.
\FOR{$k=0,1,\dots,N-1$}
    \STATE $\beta_k\leftarrow (k-k_{\rm rst}-1)/(k-k_{\rm rst}+2)$ \hfill\COMMENT{phase-local momentum coefficient}
    \STATE $y_k\leftarrow x_k+\beta_k\sqrt{s}\,v_k$
    \STATE $x_{k+1}\leftarrow y_k-s\,\nabla f(y_k)$
    \STATE $v_{k+1}\leftarrow (x_{k+1}-x_k)/\sqrt{s}$
    \IF{$\Vbar\bigl(k+1,x_{k+1},v_{k+1},x_{\rm anc};\theta_{\rm rst}\bigr)\le 0$}
        \STATE $v_{k+1}\leftarrow 0$, $\;k_{\rm rst}\leftarrow k+1$, $\;x_{\rm anc}\leftarrow x_{k+1}$
        \hfill\COMMENT{restart: zero momentum, reset phase clock and anchor}
    \ENDIF
\ENDFOR
\STATE \textbf{return} $x_N$
\end{algorithmic}
\end{algorithm}

The two standard hand-crafted restart rules of \citet{o2015adaptive}, used as baselines in \cref{sec:qp-validation,sec:qp-discrete}, are recovered by the following choices of $\Vbar$:
\begin{itemize}[leftmargin=1.5em,itemsep=0.15em]
    \item \emph{Function restart}: restart iff $f(x_{k+1})>f(x_k)$, equivalently $\Vbar^{\rm func}_k=f(x_k)-f(x_{k+1})$.
    \item \emph{Gradient restart}: restart iff $\nabla f(y_k)^\top(x_{k+1}-x_k)>0$, equivalently $\Vbar^{\rm grad}_k=-\nabla f(y_k)^\top(x_{k+1}-x_k)$.
\end{itemize}
The continuous-time analogues used in \cref{sec:qp-validation} are obtained by replacing $x_{k+1}-x_k$ by $\sqrt{s}\,\dot x(t)$ and $y_k$ by $x(t)+\sqrt{s}\,\dot x(t)$:
\begin{equation*}
    V_{\sf func}(t)=-\nabla f(x(t))^\top \dot x(t),
    \qquad
    V_{\sf grad}(t)=-\nabla f(x(t)+\sqrt{s}\,\dot x(t))^\top \dot x(t).
\end{equation*}

\subsection{Implementation via Automatic Differentiation}
\label{app:auto-diff}

\newcommand{\autograd}{\mathsf{autograd}}

We now describe how the surrogate gradients of \cref{def:disc_restart_sensitivity} can be realized as a single backward pass of a standard automatic-differentiation engine, with no manual chain-rule bookkeeping across phases. In what follows it is essential to keep two distinct gradient operators in mind:
\begin{itemize}[leftmargin=1.5em,itemsep=0.15em]
    \item $\grad(\cdot)$ denotes the \emph{surrogate gradients} introduced in \cref{def:disc_restart_sensitivity}. It is a mathematical object defined by \eqref{eq:disc_dSi_dp}--\eqref{eq:disc_dXn}, and need not be an ordinary derivative.
    \item $\autograd(\cdot)$ denotes the \emph{autodiff gradient} produced by an autodiff engine (e.g., JAX or PyTorch) using the backward rules of its computation graph, including the stop-gradient rule below.
    \item $\sgrad(\cdot)$ denotes the stop-gradient operator: it acts as the identity in the forward pass and detaches its argument from the computation graph in the backward pass.
\end{itemize}
We first construct, at each phase $i$, two auxiliary quantities---a \emph{soft restart step} $\hat S_i$ and a \emph{soft phase-end state} $\hat{x}_{i+1}$---that are numerically identical to $S_i$ and $\xbar_{i+1}$ on the forward pass, but whose autodiff gradients reproduce their symbolic sensitivities:
\begin{equation}
    \autograd(\hat S_i)=\disc{S_i},\qquad
    \autograd(\hat{x}_{i+1})=\disc{\xbar_{i+1}}.
    \label{eq:autodiff_goal}
\end{equation}
We assume inductively that the engine has been initialized so that $\autograd(\xbar_i)=\disc{\xbar_i}$ at the start of phase $i$.

\paragraph{Soft restart step and soft phase-end state.}
Building on the construction in \cref{sec:discrete}, at each phase $i$ we define the \emph{soft restart time}
\begin{equation}
    \tau_i
    :=\frac{\Fbarv(S_i;\xbar_i,\theta)}{\sgrad(\Delta_i)}
    =\frac{-\Fbarv(S_i;\xbar_i,\theta)}{\sgrad\bigl(\Fbarv(S_i-1;\xbar_i,\theta)-\Fbarv(S_i;\xbar_i,\theta)\bigr)},
    \label{eq:autodiff_tau}
\end{equation}
and the interpolation mask
\begin{equation}
    m_i:= 1-\tau_i+\sgrad(\tau_i).
    \label{eq:autodiff_mask}
\end{equation}
With this notation, the soft restart step and the soft phase-end state at phase $i$ are
\begin{equation}
    \hat S_i:= (S_i-1)+m_i,
    \qquad
    \hat{x}_{i+1}:= m_i\,\Fbarr(S_i;\xbar_i,\theta)+(1-m_i)\,\Fbarr(S_i-1;\xbar_i,\theta),
    \label{eq:autodiff_soft_objects}
\end{equation}
where, as in \cref{def:disc_restart_sensitivity}, $\Fbarr=R\circ \Fbarx$.

Both $\hat S_i$ and $\hat{x}_{i+1}$ have a transparent forward-pass behavior. Since $\tau_i$ and $\sgrad(\tau_i)$ share the same numerical value, the mask satisfies $m_i=1$ in the forward pass, and therefore
\[
    \hat S_i=S_i,
    \qquad
    \hat{x}_{i+1}=\Fbarr(S_i;\xbar_i,\theta)=\xbar_{i+1}.
\]
That is, replacing the integer-valued boundary by its soft counterpart leaves the trajectory unchanged on the forward pass, so $\hat{x}_{i+1}$ can be fed as the phase-end state into the next phase without altering the simulation.

\paragraph{Backward-pass agreement with the symbolic sensitivities.}
\begingroup
With $S_i$ held fixed and $\autograd(\xbar_i)=\disc{\xbar_i}$,
\begin{align}
 \autograd\bigl(\Fbarx(S_i;\xbar_i,\theta)\bigr)
 &=\partial_x\Fbarx(S_i;\xbar_i,\theta)\disc{\xbar_i}
    +\partial_\theta\Fbarx(S_i;\xbar_i,\theta),\notag\\
 \autograd\bigl(\Fbarv(S_i;\xbar_i,\theta)\bigr)
 &=\partial_x\Fbarv(S_i;\xbar_i,\theta)\disc{\xbar_i}
    +\partial_\theta\Fbarv(S_i;\xbar_i,\theta).
 \label{eq:autodiff_chain_rule}
\end{align}
Using $\autograd(\sgrad(\tau_i))=\autograd(\sgrad(\Delta_i))=0$,
\begin{align*}
 \autograd(m_i)
 &=-\autograd(\tau_i)
 =-\frac{\partial_\theta\Fbarv(S_i;\xbar_i,\theta)
       +\partial_x\Fbarv(S_i;\xbar_i,\theta)\disc{\xbar_i}}{\Delta_i}
 =\disc{S_i},\\
 \autograd(\hat S_i)&=\autograd(m_i)=\disc{S_i}.
\end{align*}
Since $m_i=1$ on the forward pass,
\begin{align}
 \autograd(\hat{x}_{i+1})
 &=\bigl(\Fbarr(S_i;\xbar_i,\theta)-\Fbarr(S_i-1;\xbar_i,\theta)\bigr)
       \autograd(m_i)\notag\\
 &\quad+\autograd\bigl(\Fbarr(S_i;\xbar_i,\theta)\bigr)\notag\\
 &=\bigl(\Fbarr(S_i;\xbar_i,\theta)-\Fbarr(S_i-1;\xbar_i,\theta)\bigr)
       \disc{S_i}\notag\\
 &\quad+\partial_x\Fbarr(S_i;\xbar_i,\theta)\disc{\xbar_i}
       +\partial_\theta\Fbarr(S_i;\xbar_i,\theta)\notag\\
 &=\disc{\xbar_{i+1}}.
 \label{eq:autodiff_state_backward}
\end{align}
\endgroup

\paragraph{Fixed global-time outputs.}
In addition, maintain the accumulated soft boundary $\hat N_0=0$ and $\hat N_{i+1}=\hat N_i+\hat S_i$, so that $\hat N_i=N_i$ in the forward pass and $\autograd(\hat N_i)=\disc{N_i}$. At an interior global step $n=N_i+k$, $1\leq k<S_i$, let $X_k=\Fbarx(k;\hat x_i,\theta)$ be the (uncorrected) phase-local state, with $\hat x_0=x_0$, and define the output
\begin{equation}
 \hat X(n)=X_k-\sgrad(X_k-X_{k-1})\bigl(\hat N_i-\sgrad(\hat N_i)\bigr).
 \label{eq:autodiff_global_output}
\end{equation}
The added term is zero in the forward pass, while its backward pass gives
\begin{align*}
 \autograd(\hat X(n))
 &=\partial_x\Fbarx(k)\disc{\xbar_i}+\partial_\theta\Fbarx(k)
   -(X_k-X_{k-1})\disc{N_i}\\
 &=\disc{\Xbar(n)}.
\end{align*}
Therefore, we can use $\hat X(n)$ at each fixed-global-step output or loss evaluation.

\input{revision/forward-implementation}

\section{Proofs from \cref{sec:problem_setup} and \cref{sec:discrete}}

\input{revision/restart-proofs}

\subsection{Consistency for Forward Euler Discretization}
\label{app:restart_examples}

\newcommand{\Zbar}{\bar{Z}}
\newcommand{\Wbar}{\bar{W}}

We first verify \cref{ass:disc_consistency} for the simplest case: forward Euler discretization of the generic restart-augmented ODE \eqref{eq:ODE-restart}. The argument is standard but is included here to make the use of \cref{thm:disc_cont_close} self-contained.

\paragraph{Setup.}
Consider the ODE \eqref{eq:ODE-restart}:
\begin{equation*}
    \dot X(t)=-\cAx(X(t);\theta),
    \qquad
    \dot V(t)=-\cAv(X(t),V(t);\theta),
    \qquad X(0)=x,\;V(0)=1,
\end{equation*}
and the forward Euler discretization with stepsize $h>0$:
\begin{equation}
    \Xbar_{k+1}=\Xbar_k-h\,\cAx(\Xbar_k;\theta),
    \qquad
    \Vbar_{k+1}=\Vbar_k-h\,\cAv(\Xbar_k,\Vbar_k;\theta),
    \qquad t_k=kh.
    \label{eq:app-euler-iter}
\end{equation}
The flow map $F$ and $\Fbar$ are defined accordingly.

In the following, we verify the consistency condition \eqref{eq:Fhat_consistency} and \eqref{eq:Vhat_time_consistency} separately. We pack the variable $Z=(X,V)$ and define $\cA(Z;\theta):=(\cAx(X;\theta),\cAv(X,V;\theta))$.

\paragraph{Consistency condition \eqref{eq:Fhat_consistency}}
We first prove \eqref{eq:Fhat_consistency} for this setting.

\begin{proposition}[Forward-Euler consistency I]
\label{prop:euler-consistency}
Suppose there exists $M>0$ such that $\nrm{\cA(Z;\theta)}\leq M$, $\nrm{\partial_{(Z,\theta)}\cA(Z;\theta)}\leq M$ and $\nrm{\partial_{(Z,\theta)} \cA(Z;\theta)-\partial_{(Z,\theta)} \cA(Z';\theta)}\leq M\nrm{Z-Z'}$ for all $Z,Z'$. Then there exists $C=C(M,T)>0$ such that, for any stepsize $h>0$ and any integer $k$ with $t_k\leq T$,
\begin{align}
    \nrm{\Fbar(k;x,\theta)-F(t_k;x,\theta)}&\leq Ch, \label{eq:app-euler-state-consistency} \\
    \nrm{\partial_{(x,\theta)}\Fbar(k;x,\theta)-\partial_{(x,\theta)}F(t_k;x,\theta)}
    &\leq Ch. \label{eq:app-euler-sensitivity-consistency}
\end{align}
\end{proposition}

\begingroup
\begin{proof}
Let $Z=(X,V)$, $\dot Z=-\cA(Z;\theta)$, $\Zbar_{k+1}=\Zbar_k-h\cA(\Zbar_k;\theta)$, and $Z(0)=\Zbar_0=(x,1)$.

\paragraph{State error.}
For $e_k=\|Z(t_k)-\Zbar_k\|$,
\begin{gather*}
 \ddot Z=\partial_Z\cA(Z;\theta)\cA(Z;\theta),\qquad
 \|\ddot Z\|_\infty\leq M^2,\\
 r_k:=Z(t_{k+1})-Z(t_k)+h\cA(Z(t_k);\theta)
 =\int_{t_k}^{t_{k+1}}(t_{k+1}-s)\ddot Z(s)\,ds,
 \qquad \|r_k\|\leq M^2h^2/2,\\
 e_{k+1}\leq(1+Mh)e_k+M^2h^2/2,\qquad e_0=0,\\
 e_k\leq\frac{M^2h^2}{2}\sum_{j=0}^{k-1}(1+Mh)^j
 \leq\frac{Mh}{2}(e^{MT}-1).
\end{gather*}

\paragraph{Sensitivity error.}
Set $W=D_{(x,\theta)}Z$, $\Wbar_k=D_{(x,\theta)}\Zbar_k$, and
$B(Z)=\partial_Z\cA(Z;\theta)$, $C(Z)=[0\ \partial_\theta\cA(Z;\theta)]$. Then
\begin{align*}
 \dot W&=-B(Z)W-C(Z),\\
 \Wbar_{k+1}&=\Wbar_k-hB(\Zbar_k)\Wbar_k-hC(\Zbar_k),\qquad
 W(0)=\Wbar_0.
\end{align*}
\begin{gather*}
 \sup_t\|W(t)\|+\sup_k\|\Wbar_k\|\leq C(M,T),\\
 \|Z(t)-Z(s)\|+\|W(t)-W(s)\|\leq C|t-s|.
\end{gather*}
\begin{align*}
 \|\dot W(t)-\dot W(s)\|
 &\leq\|B(Z(t))-B(Z(s))\|\,\|W(t)\|\\
 &\quad+\|B(Z(s))\|\,\|W(t)-W(s)\|
       +\|C(Z(t))-C(Z(s))\|\leq C_1|t-s|,\\
 r_k^W&:=W(t_{k+1})-W(t_k)-h\dot W(t_k)
       =\int_{t_k}^{t_{k+1}}[\dot W(s)-\dot W(t_k)]\,ds,\\
 \|r_k^W\|&\leq C_1h^2/2.
\end{align*}
Thus, with $\tilde e_k=\|W(t_k)-\Wbar_k\|$,
\begin{gather*}
 \tilde e_{k+1}\leq(1+Mh)\tilde e_k
       +Mh(\|W(t_k)\|+1)e_k+C_1h^2/2
       \leq(1+Mh)\tilde e_k+Ch^2,\qquad \tilde e_0=0,\\
 \tilde e_k\leq Ch^2\sum_{j=0}^{k-1}(1+Mh)^j\leq Ch.
\end{gather*}
\end{proof}
\endgroup

\paragraph{Consistency condition \eqref{eq:Vhat_time_consistency}} Next, we prove \eqref{eq:Vhat_time_consistency}.

\begin{proposition}[Forward-Euler consistency II]
\label{prop:euler-consistency-ii}
Under the assumptions of \cref{prop:euler-consistency}, there exists $C'=C'(M,T)>0$ such that, for any stepsize $h>0$ and any integer $k\geq0$ with $t_{k+1}\leq T$,
\begin{equation}
    \nrm*{\frac{\Fbar(k+1;x,\theta)-\Fbar(k;x,\theta)}{h}-\partial_t F(t_k;x,\theta)}
    \leq C'h.
    \label{eq:app-euler-time-derivative-consistency}
\end{equation}
\end{proposition}

\begingroup
\begin{proof}
For $0\leq t_k<t_{k+1}\leq T$, \eqref{eq:app-euler-iter} and \cref{prop:euler-consistency} give
\begin{align*}
 \frac{\Fbar(k+1;x,\theta)-\Fbar(k;x,\theta)}h
   &=-\cA(\Zbar_k;\theta),\\
 \partial_tF(t_k;x,\theta)&=-\cA(Z(t_k);\theta),\\
 \|\cA(\Zbar_k;\theta)-\cA(Z(t_k);\theta)\|
 &\leq M\|\Zbar_k-Z(t_k)\|\leq \frac{M^2h}{2}(e^{MT}-1)=C'h.
\end{align*}
\end{proof}
\endgroup

\input{revision/nag-consistency}

\section{Experiment Details for QP}\label{app:QP}

We provide details for the experiments in
\cref{sec:qp-validation}, \cref{sec:disc_validation}, and \cref{sec:qp-discrete}.

\paragraph{Problem instances.}
With a given dimension $d$ and condition number $\kappa$, we generate QP problems of the form \eqref{eq:QP} as \footnote{We remark that it is without loss of generality to assume the matrix $A$ in \eqref{eq:QP} is diagonal as the NAG updates are rotation-invariant.}
\begin{equation}
    f(x)=\tfrac12 (x-x^\star)^\top Q(x-x^\star),
    \qquad
    Q=\operatorname{diag}(\lambda_1,\ldots,\lambda_d),
    \label{eq:app-qp-objective}
\end{equation}
where the diagonal entries are placed on a geometric grid,
\begin{equation}
    \lambda_i=\kappa^{(i-1)/(d-1)},\qquad i=1,\dots,d,
    \label{eq:app-qp-eigvals}
\end{equation}
so that the smallest and largest eigenvalues are $\mu=1$ and $L=\kappa=80$. This corresponds to setting $A=Q$, $b=Qx^\star$ and $c=\frac{1}{2}(x^\star)^\top Qx^\star$.

The minimizer $x^\star$ is sampled from the standard Gaussian distribution with random seed $0$. The initial point is sampled as $x_0 \sim \mathrm{Unif}([-20,20]^d)$.

\subsection{Details of \cref{sec:qp-validation}}
\label{app:QP-ODE}

\paragraph{ODE dynamics.}
Within each restart phase, the continuous-time dynamics are the high-resolution ODE~\eqref{eq:NAG-C-ODE}, written in joint state $(x(t),v(t))$ with $v=\dot x$:
\begin{equation}\label{eq:NAG-C-ODE-app}
    \textstyle
        \dot{x}(t)=v(t), \qquad
        \dot{v}(t)=-\dfrac{3}{t+\tau}v(t)-\sqrt{s}\,\nabla^2 f(x(t))\,v(t)-\left(1+\dfrac{3\sqrt{s}}{2(t+\tau)}\right)\nabla f(x(t)).
\end{equation}
with $\tau=3\sqrt{s}/2$, $x(0)=x_0$ and $\dot{x}(0)=-\sqrt{s}\nabla f(x_0)$.
The reference stepsize is $s=1/L=0.0125$ and the terminal time is $T=K\sqrt{s}$ with $K=200$. Trajectories are integrated with a fourth-order Runge--Kutta method at fixed step $\Delta t=5\times 10^{-3}$. When a restart is triggered, the phase is reset by $V\leftarrow 1$  and the current state is stored as the new phase anchor.

As a reference-only ceiling, in \cref{fig:qp-validation} we also report the high-resolution accelerated ODE for $\mu$-strongly-convex objectives of \citet{shi2022understanding}:
\begin{equation}\label{eq:NAG-SC-ODE-app}
    \ddot{x}(t)+2\sqrt{\mu}\,\dot{x}(t)+\sqrt{s}\,\nabla^2 f(x(t))\,\dot{x}(t)+\bigl(1+\sqrt{\mu s}\bigr)\nabla f(x(t))=0,
\end{equation}
with $x(0)=x_0$ and $\dot{x}(0)=-\tfrac{2\sqrt{s}}{1+\sqrt{\mu s}}\nabla f(x_0)$. Unlike the ODE~\eqref{eq:NAG-C-ODE}, this dynamics requires the strong-convexity parameter $\mu$ explicitly, and is therefore not a practical baseline; it is included here only as a reference that adaptive restart aims to recover without knowledge of $\mu$.

\paragraph{Network architecture and training.}
The learnable score $V(X;\theta)=\MLP(\phi(X,t);\theta)$ is an MLP with embedding map $\phi:\RR^d\to \RR^5$ defined as
\begin{equation}
    \phi([x;v],t)
    =\bigl(f(x),\ \nrm{\nabla f(x)},\ \nrm{x-x_{\rm anc}},\ \sqrt{s}\,\nrm{v},\ 1+t/\sqrt{s}\bigr),
    \label{eq:app-qp-features}
\end{equation}
where $x_{\rm anc}$ is the starting point of the current phase.
The MLP has input dimension $5$, a single hidden layer of width $16$ with $\tanh$ activations, and a scalar output logit, for a total of $113$ trainable parameters.

Training proceeds in two stages. We first \emph{warm-start} the network on a dataset of feature--label pairs $\mathcal{D}=\{(\phi_i,y_i)\}_{i=1}^{M}$ with $M=800$ examples collected from trajectories driven by the function-restart ODE; the warm-start is fitted by minimizing the binary cross-entropy with positive-class reweighting to compensate for the rarity of restart labels. Starting from the warm-started parameters, we then \emph{fine-tune} the network using the surrogate gradients of \cref{def:disc_restart_sensitivity} on the loss \eqref{eq:qp-loss} for $20$ epochs, with the AMSGrad optimizer at learning rate $10^{-2}$.

\paragraph{Restart policies.}
\Cref{fig:qp-validation} compares six trajectories on the test instance:
\begin{itemize}[leftmargin=1.5em,itemsep=0.15em]
    \item \emph{ODE no restart}: the ODE~\eqref{eq:NAG-C-ODE-app} with no restart applied.
    \item \emph{Function-restart ODE}: restart triggered by the score $V_{\sf func}(t)=-\nabla f(X)^\top\dot X$, the continuous analogue of the discrete rule ``restart iff $f(x_{k+1})>f(x_k)$''.
    \item \emph{Gradient-restart ODE}: restart triggered by the score $V_{\sf grad}(t)=-\nabla f(X+\sqrt{s}\,\dot X)^\top\dot X$, the continuous analogue of the discrete rule ``restart iff $\nabla f(y_k)^\top(x_{k+1}-x_k)>0$''.
    \item \emph{Periodic-restart ODE}: restart triggered at $t=\tau,2\tau,\cdots$ for a fixed choice of restart length $\tau$.
    \item \emph{Oracle SC-NAG ODE}: the strongly-convex accelerated ODE that uses the true strong-convexity constant $\mu$; this is a reference-only ceiling, since $\mu$ is not available in practice.
    \item \emph{Learned restart}: the trained policy.
\end{itemize}
For both handcrafted scores, a restart is triggered when the corresponding score crosses from nonnegative to negative between two consecutive RK4 states (i.e., $V\le 0$); the precise crossing time is approximated by linear interpolation within the step.

\paragraph{Final results on the test instance.}
\Cref{tab:qp-final-gaps} reports the terminal objective gaps at $t=T$ for the trajectories plotted in \cref{fig:qp-validation}.

\begin{table}[ht]
\centering
\caption{Terminal objective gaps $f(x(T))-\min f$ on the held-out test initialization ($d=30$, $\kappa=80$, $T\approx 22.36$). The learned restart policy improves over both handcrafted scores by several orders of magnitude and approaches the oracle strongly-convex curve.}
\label{tab:qp-final-gaps}
\begin{tabular}{lc}
\toprule
Method & Terminal gap $f(x(T))-\min f$ \\
\midrule
ODE no restart            & $4.99\times 10^{-5}$ \\
Function-restart ODE       & $2.12\times 10^{-11}$ \\
Gradient-restart ODE       & $7.02\times 10^{-12}$ \\
Learned restart            & $4.75\times 10^{-14}$ \\
Oracle SC-NAG ODE (reference) & $1.28\times 10^{-17}$ \\
\bottomrule
\end{tabular}
\end{table}

\subsection{Validation of the Approximation Accuracy}
\label{sec:disc_validation}

\begin{figure}[t]
    \centering
    \includegraphics[width=0.8\textwidth]{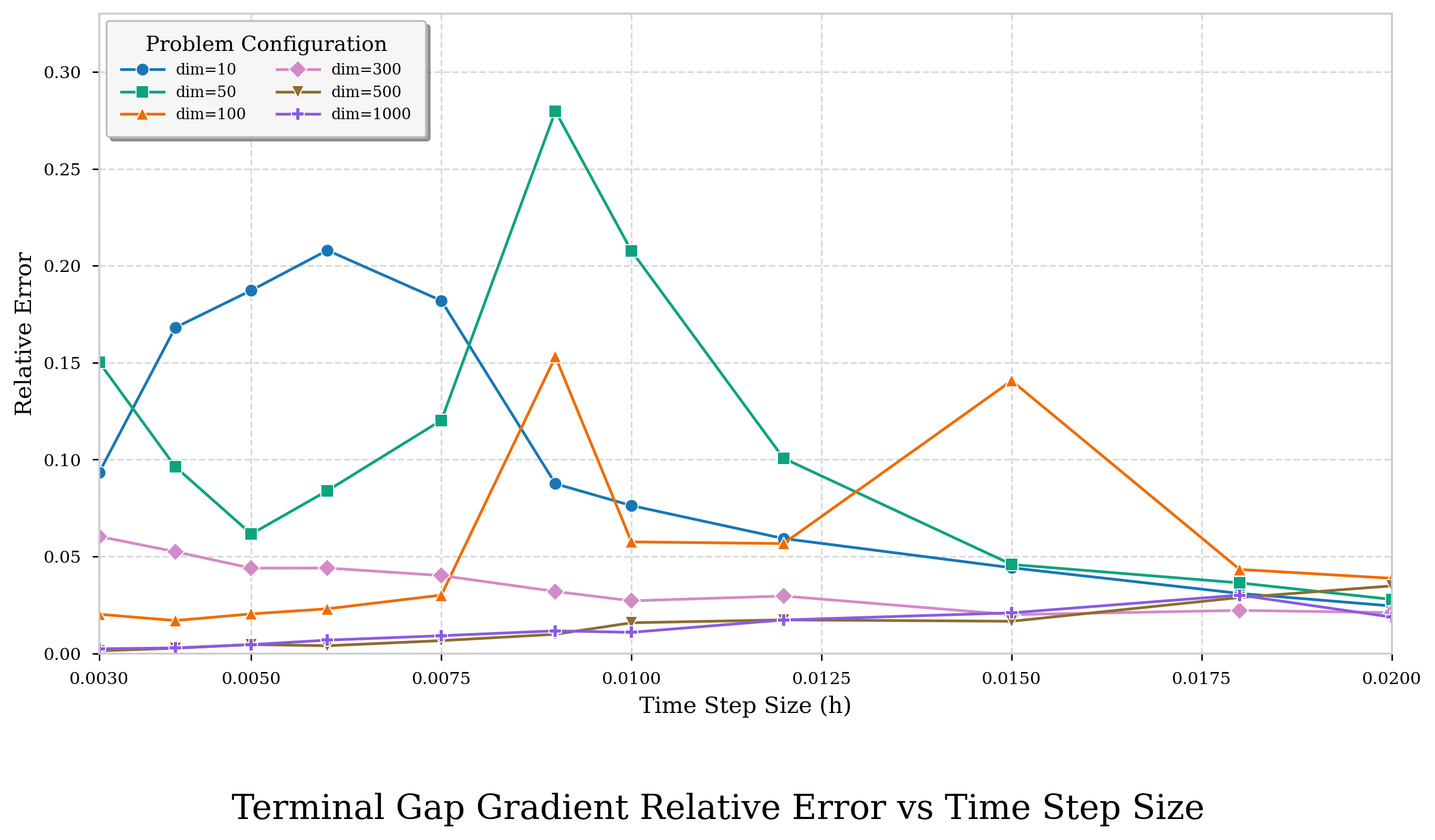}
    \caption{Relative error between the discrete surrogate gradient $\grad(f(X(T)))$ and the continuous-time gradient $\partial_\theta f(X(T))$, plotted as a function of the stepsize $h$ (with $s=h^2$) for dimensions $d$ ranging from $10$ to $1000$. The error decreases as $h\to 0$ at a rate consistent with \cref{thm:disc_cont_close} and remains stable across dimensions. Details are in \cref{app:QP-ODE}.
    }
    \label{fig:disc_validation}
\end{figure}

We now empirically validate the surrogate gradient of \cref{def:disc_restart_sensitivity} against its continuous-time counterpart in a setting where both can be computed accurately. We reuse the QP problem family of \cref{sec:qp-validation} together with a simple parameterized restart score $V(\cdot;\theta)$. We compare:
\begin{itemize}[leftmargin=1.5em,itemsep=0.15em]
    \item Continuous-time dynamics $(X(t))$ driven by ODE~\eqref{eq:NAG-C-ODE} with gradients computed via \cref{def:continuous_restarted_trajectory}.
    \item Discrete-time dynamics $(\Xbar(n))$ driven by NAG iteration with stepsize $s$ (detailed in \cref{app:algorithms_with_restarts}) with gradients computed via \cref{def:disc_restart_sensitivity} (via the autodiff implementation of \cref{app:auto-diff}).
\end{itemize}

We evaluate the average relative error
\begin{equation}
    \mathrm{Err}_{\rm grad}
    =\frac{\nrm{\disc{\Xbar(n)}-\partial_\theta X(t_n)}_2}{\nrm{\partial_\theta X(t_n)}_2+\nrm{\disc{\Xbar(n)}}_2},
\end{equation}
across a range of stepsizes $s$ and dimensions $d$, and report them in \cref{fig:disc_validation}. As predicted by \cref{thm:disc_cont_close} together with the consistency analysis of \cref{app:nag_consistency}, both errors decrease as $s\to 0$ and remain stable across dimensions, confirming that the surrogate gradient faithfully approximates its continuous counterpart at the same order as the underlying discretization error.

\subsection{Details for \cref{sec:qp-discrete}}\label{app:qp-discrete}

The base optimizer is the NAG iteration of \cref{app:algorithms_with_restarts} with step size $\eta = 1/\lambda_{\max}(H)$.
The learned restart rule is a two-layer $\tanh$ MLP with embedding dimension $4$ and hidden dimension $32$
and input features
\begin{equation}
    \phi_k =
    \bigl(f(x_{k+1}),\;
    \nrm{x_{k+1}-\xbar_i}_2,\;
    1+(k-N_i)/\sqrt{s},\;
    \cos\angle(x_{k+1}-x_k,\;x_k-x_{k-1})\bigr),
    \label{eq:app-qp-disc-features}
\end{equation}
where $\xbar_i$ is the last restart state.

Training uses $20$-dimensional problems with condition numbers $\kappa\in\{180,240,320\}$ and the two-stage procedure of \cref{sec:qp-discrete}: warm-start from function-restart teacher labels, followed by end-to-end fine-tuning on a ratio objective that normalizes the learned policy's terminal gap by the best handcrafted baseline. The mean objective gap along the trajectory on $d=20$, $400$, and $1000$ is reported in \cref{fig:qp-id-main} in the main text.

\begin{figure}[t]
    \centering
    \includegraphics[width=1.0\linewidth]{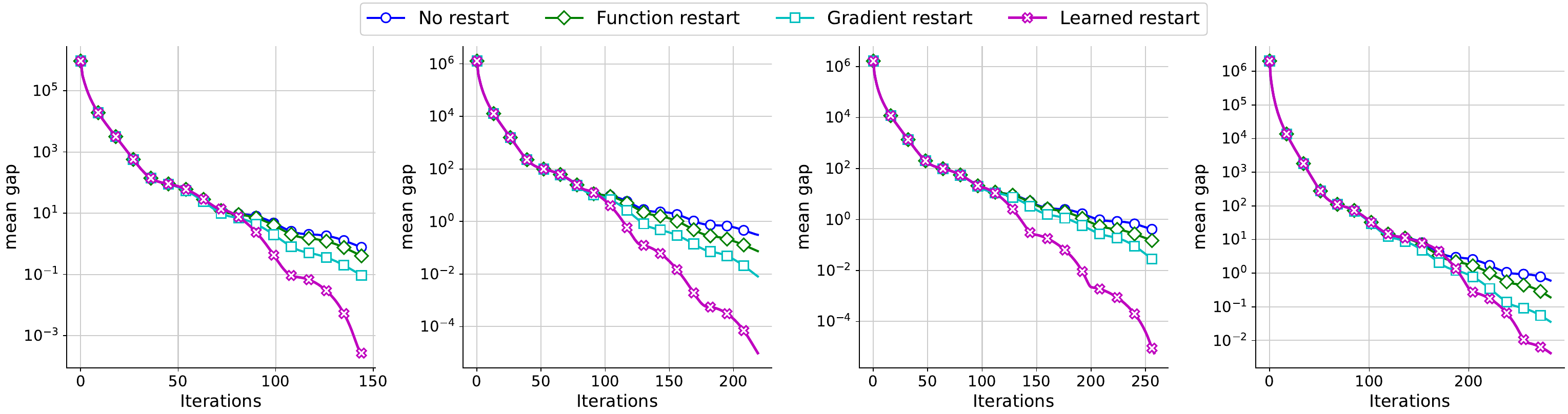}
    \includegraphics[width=1.0\linewidth]{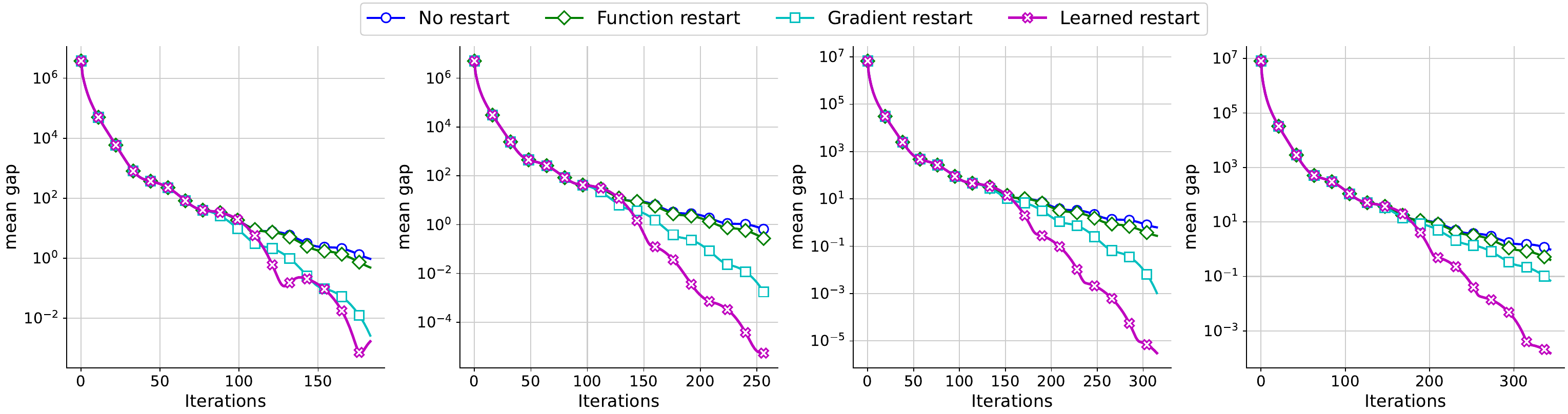}

    \caption{Out-of-distribution transfer to unseen condition numbers. The policy is trained on $\kappa\in\{180,240,320\}$ at $d=20$ and evaluated zero-shot on $d=500$ (\emph{top}) and $d=2000$ (\emph{bottom}) with $\kappa\in\{140,200,280,360\}$. The learned policy transfers across both dimensions and condition numbers not seen during training.}
    \label{fig:qp-id-app}
\end{figure}

\paragraph{Additional out-of-distribution evaluation.}
To further probe generalization, we evaluate the same policy on condition numbers \emph{not seen during training}: $\kappa\in\{140,200,280,360\}$. \Cref{fig:qp-id-app} shows results on $d=500$ and $d=2000$; the learned policy continues to outperform both handcrafted baselines on these unseen $\kappa$ values.

\section{Experiment Details for Real-Data Logistic Regression}
\label{app:logreg}

This appendix collects the details deferred from \cref{sec:logreg}: the construction of the evaluation instances, the exact computation of $f^\star$, the training distribution and protocol of the transferred policy, and the evaluation protocol.

\paragraph{Problem instances.}
We use the five binary-classification datasets \texttt{a9a}, \texttt{w8a}, \texttt{ijcnn1}, \texttt{covtype} (binarized) and \texttt{phishing} from the LIBSVM collection~\citep{libsvm}, with raw dimensions $123$, $300$, $22$, $54$ and $68$ respectively. Each \emph{instance} is generated from a dataset $(a_i,y_i)_{i=1}^{M}$, $y_i\in\{-1,+1\}$, by the following deterministic-given-seed procedure.
\begin{enumerate}[leftmargin=1.5em,itemsep=0.15em]
    \item Draw $m=1000$ rows without replacement and standardize each raw feature column to zero mean and unit variance.
    \item Draw random Fourier features~\citep{rahimi2007random} $z(a)=\sqrt{2/D}\,\cos(Wa+b)$ with $D=2000$, $W_{ij}\sim\cN(0,\sigma^{-2})$ and $b_i\sim\mathrm{Unif}[0,2\pi]$, where the bandwidth $\sigma^2$ is set by the median heuristic on a $300$-point subsample of pairwise squared distances.
    \item Rescale the resulting design matrix $Z\in\RR^{m\times D}$ to unit spectral norm, so that the data-term smoothness constant $L_0=0.25/m$ is shared across instances (using $\sup_t \psi''(t)\le 1/4$ for the logistic loss $\psi(t)=\log(1+e^{-t})$).
    \item Set $\lambda=\lambda_{\rm rel}L_0$ and $x_0\sim\cN(0,I_D)$.
\end{enumerate}
The resulting objective is the $\ell_2$-regularized logistic regression
\begin{equation}
    \min_{x\in\RR^D}\;
    f(x)=\frac{1}{m}\sum_{i=1}^{m}\log\bigl(1+e^{-y_i\tri{z_i,x}}\bigr)
    +\frac{\lambda}{2}\nrm{x}^2,
    \qquad y_i\in\{-1,+1\},
    \label{eq:logreg}
\end{equation}
where $\lambda=\lambda_{\rm rel}L_0$ with $L_0=0.25/m$ the smoothness constant of the data term. Since $\nabla^2 f\preceq \tfrac{1}{4m}Z^\top Z+\lambda I\preceq(L_0+\lambda)I$ and $\nabla^2 f\succeq\lambda I$, the instance is $(L_0+\lambda)$-smooth and $\lambda$-strongly convex, giving the condition number bound $\kappa_{\rm eff}\leq(L_0+\lambda)/\lambda=1+1/\lambda_{\rm rel}$.

\paragraph{Exact minimizer.}
Unlike the QP family, \eqref{eq:logreg} has no closed-form minimizer. We compute $x^\star$ by damped Newton with Armijo backtracking in the $D$-dimensional primal, started from the origin and iterated until $\nrm{\nabla f}_\infty\le 10^{-15}\cdot\max\{1,\nrm{\nabla f(0)}_\infty\}$; the achieved residual is recorded per instance and is of order $10^{-18}$ in all runs. The reported gaps $f(x_N)-f^\star$ are therefore exact to machine precision and involve no long reference run, which matters here because the gaps of interest are as small as $10^{-6}$.

\paragraph{Training distribution and policy.}
The policy has the same architecture family as in \cref{sec:qp-discrete} and is trained exclusively on synthetic quadratics of the form~\eqref{eq:app-qp-objective}; no real data and no logistic objective is seen during training. The training distribution draws, per problem, a dimension $d\in\{10,20,40,100\}$, a condition number $\kappa\in\{120,180,240,320,420\}$, a spectrum shape in $\{$geometric, power, log-uniform$\}$, and an initialization scale $x_0\sim\mathrm{Unif}([-c,c]^d)$ with $c\in\{5,20,50\}$; each problem is a batch of $8$ instances run for $N_{\rm train}=150$ iterations.

For transfer across problem scales the feature map~\eqref{eq:app-qp-disc-features} is replaced by a log-scaled extension: magnitudes enter through $\log_{10}$ and all remaining coordinates are dimensionless ratios,
\begin{equation}
 \begin{aligned}
    \phi_k=\Bigl(&
    \tfrac{1}{10}\log_{10} f(x_{k+1}),\;
    \tfrac{1}{10}\log_{10}\nrm{x_{k+1}-\xbar_i},\;
    \log_{10}(1+k-N_i),\;
    \\ &
    \cos\angle(x_{k+1}-x_k,x_k-x_{k-1}),\;
    r_k,\;
    c_k,\;
    p_k
    \Bigr),
 \end{aligned}
    \label{eq:app-logreg-features}
\end{equation}
where $r_k=\frac{f(x_k)-f(x_{k+1})}{f(x_k)+f(x_{k+1})}$ is the normalized one-step decrease (the scale-free analogue of the function-restart signal), $c_k=\frac{\tri{\nabla f(x_{k+1}),v_{k+1}}}{\nrm{\nabla f(x_{k+1})}\nrm{v_{k+1}}}$ is the gradient--momentum alignment (the scale-free analogue of the gradient-restart signal), and $p_k=\frac{1}{10}\log_{10}\bigl(f(x_{k+1})/f(\xbar_i)\bigr)$ measures within-phase progress relative to the phase anchor $\xbar_i$. No coordinate depends explicitly on $d$ or $m$, so the same architecture can be evaluated at higher dimension $D=2000$ after training at $d\le 100$. The features are not invariant to objective rescaling: replacing $f$ by $af$, $a>0$, shifts the first coordinate by $\tfrac1{10}\log_{10}a$. The network is a two-layer $\tanh$ MLP of width $32$ with a scalar output logit ($7$-$32$-$1$, $289$ trainable parameters), whose output bias is initialized to $+0.5$ so that the initial policy does not restart.

\paragraph{Training protocol.}
Training follows the same two stages as \cref{sec:qp-discrete}. In the \emph{warm-start} stage we roll out the function-restart rule on $24$ sampled training problems and record every $(\phi_k,\mathbbm{1}\{\text{restart}\})$ pair along the trajectories, giving $24\times 8\times 150=28{,}800$ examples; the network is fitted for $300$ full-batch Adam steps at learning rate $10^{-2}$ on the binary cross-entropy with restart logit $-V(\phi;\theta)$ and positive-class reweighting to compensate for the rarity of restart labels (converged loss $0.062$--$0.070$ across seeds). Because a restart is a hard threshold event whose straight-through gradient vanishes identically, this cloned policy is also the strongest derivative-free trainable baseline available.

In the \emph{fine-tuning} stage the warm-started network is trained end-to-end through the surrogate gradient of \cref{def:disc_restart_sensitivity} for $30$ epochs of $8$ freshly sampled problems each, minimizing the clipped log-ratio of the terminal gap to the best handcrafted baseline on the same problem,
\begin{equation}
 \begin{aligned}
    J(\theta)&=\EE\Bigl[\max\bigl\{\log\bigl(f(x_{N_{\rm train}};\theta)-f^\star\bigr)-\log g_{\rm base},\,-3\log 10\bigr\}\Bigr],\\
    g_{\rm base}&=\min_{\rm base}\bigl(f(x_{N_{\rm train}}^{\rm base})-f^\star\bigr),
 \end{aligned}
    \label{eq:app-logreg-loss}
\end{equation}
where the minimum is over $\{$no restart, function restart, gradient restart$\}$ and the floor at $-3\log 10$ prevents a single easy problem from dominating the batch. We use AMSGrad at learning rate $10^{-3}$ with gradient-norm clipping at $5$, and select the checkpoint attaining the best value of~\eqref{eq:app-logreg-loss} on a held-out set of $16$ validation problems drawn from the same synthetic distribution. The entire two-stage procedure takes about two CPU-minutes per seed. We repeat it for $5$ independent seeds and report all of them; no seed selection is performed anywhere in \cref{sec:logreg}.

\paragraph{Evaluation protocol.}
All methods are run on the same instance, from the same $x_0$, with the same stepsize $s=1/(L_0+\lambda)$ and the same horizon $N$; the only difference is the restart rule. We evaluate $6$ instances per configuration and all $5$ training seeds, giving $30$ paired $(\text{instance},\text{seed})$ comparisons per cell and $300$ in total across the ten configurations. Per-configuration numbers are geometric means of the terminal gap over instances (and additionally over seeds for the learned policies). Because different datasets have gaps of different magnitude, aggregate improvements are reported as geometric means of the per-configuration ratios. We compare against the instance-wise \emph{envelope} of the three handcrafted rules, $g_{\rm env}=\min\{g_{\rm none},g_{\rm func},g_{\rm grad}\}$, which is a strictly stronger competitor than any fixed single rule since it is allowed to select the best criterion separately on each instance.

\begin{table}[ht]
\centering\footnotesize\setlength{\tabcolsep}{6pt}
\caption{Terminal gap $f(x_N)-f^\star$ on real-data logistic regression~\eqref{eq:logreg} ($m=1000$, $D=2000$). Each dataset occupies two rows, one per regime: the \textbf{first row} is the regularization level $\lambda_{\rm rel}=10^{-4}$ with iteration budget $N=500$, and the \textbf{second row} is $\lambda_{\rm rel}=3\cdot 10^{-5}$ with $N=1000$; recall $\lambda=\lambda_{\rm rel}L_0$.}
\label{tab:logreg-main}
\begin{tabular}{lrrrr}
\toprule
Dataset & No restart & Function & Gradient & Learned\\
\midrule
\multirow{2}{*}{\texttt{a9a}}
   & 2.89e-04 & 1.16e-04 & 1.13e-04 & \textbf{8.19e-05}\\
 & 1.74e-04 & 1.74e-04 & 1.74e-04 & \textbf{2.33e-05}\\
\addlinespace[2pt]
\multirow{2}{*}{\texttt{w8a}}
   & 3.15e-05 & 2.14e-05 & 2.08e-05 & \textbf{7.82e-06}\\
 & 1.30e-05 & 3.51e-06 & 3.45e-06 & \textbf{1.55e-06}\\
\addlinespace[2pt]
\multirow{2}{*}{\texttt{ijcnn1}}
   & 1.45e-04 & 9.52e-05 & 7.73e-05 & \textbf{4.35e-05}\\
 & 7.48e-05 & 5.02e-05 & 4.93e-05 & \textbf{1.00e-05}\\
\addlinespace[2pt]
\multirow{2}{*}{\texttt{covtype}}
   & 2.64e-04 & 1.66e-04 & 1.63e-04 & \textbf{7.41e-05}\\
 & 1.71e-04 & 6.90e-05 & 6.67e-05 & \textbf{2.81e-05}\\
\addlinespace[2pt]
\multirow{2}{*}{\texttt{phishing}}
   & 1.06e-04 & 4.59e-05 & 4.54e-05 & \textbf{3.03e-05}\\
 & 6.71e-05 & 5.06e-05 & 5.04e-05 & \textbf{1.07e-05}\\
\bottomrule
\end{tabular}
\end{table}

\paragraph{Full results.}
\Cref{tab:logreg-main} reports the terminal gaps of the learned policy against the three handcrafted rules.

\section{Experiment Details for Real-Measurement Computed Tomography}
\label{app:ct}

This appendix collects the details deferred from \cref{sec:ct}, the second zero-shot transfer experiment, complementary to \cref{app:logreg}: the same synthetic-QP training pipeline is applied, without retraining, to ridge reconstruction from \emph{physically measured} X-ray tomography data.
\paragraph{Problem instances.}
We use three open tomographic datasets released by the Finnish Inverse Problems Society: a walnut scanned at an $82\times 82$ reconstruction grid~\citep{walnutdata}, and the carved-cheese phantom scanned at a $128\times128$ grid in both a full-angle and a limited-angle configuration~\citep{cheesedata}. Each dataset supplies a sparse fan-beam system matrix $A\in\RR^{m\times d}$ and the measured sinogram $b\in\RR^m$; no forward projection is simulated and no synthetic noise is added. We denote the three problems \texttt{walnut82}, \texttt{cheeseF45} and \texttt{cheeseL45}. Their dimensions are $m\times d=9840\times 6724$ with $9.70\cdot 10^5$ nonzeros ($1.47\%$ dense, $120$ projection angles $\times$ $82$ detector bins) for \texttt{walnut82}, and $44505\times 16384$ with $8.54\cdot 10^{6}$ nonzeros ($1.17\%$ dense, $45$ projection angles $\times$ $989$ detector bins) for both cheese problems. The two cheese operators have identical size and differ only in angular coverage: \texttt{cheeseF45} distributes its $45$ angles over the full circle, whereas \texttt{cheeseL45} confines them to a restricted arc, which makes the limited-angle operator substantially more ill-posed. This is a standard difficulty axis in tomography and is the reason both configurations are reported.

The reconstruction problem is the ridge least-squares objective
\begin{equation}
    \min_{x\in\RR^{d}}\;
    f(x)=\frac{1}{2m}\nrm{Ax-b}^2+\frac{\lambda}{2}\nrm{x}^2,
    \qquad
    \lambda=\lambda_{\rm rel}L_0,
    \quad
    L_0=\sigma_{\max}^2(A)/m,
    \label{eq:ct}
\end{equation}
where $L_0$ is computed by $300$ power iterations on $A^\top A$ and cached. Since $\nabla^2 f=\frac{1}{m}A^\top A+\lambda I$, the instance is $(L_0+\lambda)$-smooth and $\lambda$-strongly convex, and the Hessian condition number is $\kappa_{\rm eff}=(L_0+\lambda)/(\lambda_{\min}(A^\top A)/m+\lambda)\leq1+1/\lambda_{\rm rel}$. The three datasets differ by an order of magnitude in the scale of the data themselves --- $L_0=0.254$ for \texttt{walnut82} against $10.8$ and $11.4$ for \texttt{cheeseF45} and \texttt{cheeseL45}, with sinogram entries in $[0,0.92]$ versus $[0.013,2.45]$ --- so these datasets also test transfer across objective scales.

A \emph{configuration} is a triple (dataset, $\lambda_{\rm rel}$, $N$), and each configuration contains $8$ instances that share $(A,b,\lambda)$ and differ only in the initial point,
$x_0=c_j\cdot\frac{\nrm{x^\star}}{\sqrt{d}}\cdot\xi_j$ with $\xi_j\sim\cN(0,I_d)$ drawn from a fixed seed and $c_j$ alternating over $\{0.5,2.0\}$ across $j=0,\dots,7$, so that the two groups have different starting magnitudes. This makes the $8$ instances a controlled sweep over initialization distance rather than $8$ i.i.d.\ repetitions.

\paragraph{Exact minimizer.}
Unlike \eqref{eq:logreg}, \eqref{eq:ct} is a quadratic, so $x^\star$ solves the normal equations $\bigl(\tfrac1m A^\top A+\lambda I\bigr)x=\tfrac1m A^\top b$. We solve them by conjugate gradients to machine precision ($\texttt{rtol}=10^{-16}$) and certify the result by the strong-convexity bound $f(x)-f^\star\le\nrm{\nabla f(x)}^2/(2\lambda)$, which is evaluated at the returned iterate and asserted to be below $10^{-13}f^\star$; the measured certificates lie between $6.7\cdot10^{-29}$ and $8.6\cdot10^{-27}$, i.e.\ at least $14$ orders of magnitude below the smallest gap appearing in \cref{tab:ct-main}. The reported gaps therefore involve no long reference run, which matters here because the smallest entry of the table is of order $10^{-13}$. As a sanity check on the reconstructions themselves, the relative data residual $\nrm{Ax^\star-b}/\nrm{b}$ at $\lambda_{\rm rel}=3\cdot10^{-4}$ is $2.8\%$, $4.4\%$ and $7.2\%$ on \texttt{walnut82}, \texttt{cheeseL45} and \texttt{cheeseF45}.

\paragraph{Training distribution and policy.}
The policy is trained exclusively on synthetic diagonal quadratics of the form~\eqref{eq:app-qp-objective}; no tomographic data, no measured operator and no sparse problem is seen during training. Per problem the training distribution draws a dimension $d\in\{50,200,500\}$, a condition number $\kappa\in\{300,10^3,3\cdot10^3,10^4\}$, a spectrum shape in $\{$geometric, power, log-random, bulk-tail$\}$, and an initialization scale $x_0\sim\mathrm{Unif}([-c,c]^d)$ with $c\in\{5,20\}$; each problem is a batch of $6$ instances. The feature map is the log-scaled map~\eqref{eq:app-logreg-features} and the network is the same two-layer $\tanh$ MLP as in \cref{app:logreg} ($7$-$32$-$1$, $289$ parameters, output bias initialized to $+0.5$ so the initial policy never restarts); all arithmetic is in float64.

\paragraph{Training protocol.}
Training follows the two stages of \cref{app:logreg}. The \emph{warm-start} stage rolls out the function-restart rule on $24$ sampled training problems and fits the network for $300$ full-batch Adam steps at learning rate $10^{-2}$ on the positively reweighted binary cross-entropy with restart logit $-V(\phi;\theta)$. The \emph{fine-tuning} stage then trains end-to-end through the surrogate gradient of \cref{def:disc_restart_sensitivity} for $40$ epochs of $5$ freshly sampled problems each, minimizing the clipped log-ratio objective~\eqref{eq:app-logreg-loss} with AMSGrad at learning rate $10^{-3}$ and gradient-norm clipping at $5$; the checkpoint with the best score on $8$ held-out synthetic validation problems is retained. No tomographic result is used at any point to select the checkpoint.

\paragraph{Evaluation protocol.}
All methods are run on the same instance, from the same $x_0$, with the same stepsize $s=1/(L_0+\lambda)$, the same NAG momentum $\beta_k=(k-1)/(k+2)$ and the same horizon $N=800$; the only difference is the restart rule. Per-configuration numbers are geometric means of the terminal gap $f(x_N)-f^\star$ over the $8$ instances. As in \cref{app:logreg} we additionally compare against the instance-wise \emph{envelope} $g_{\rm env}=\min\{g_{\rm none},g_{\rm func},g_{\rm grad}\}$, a strictly stronger competitor than any single handcrafted rule.

\paragraph{Results.}
\Cref{tab:ct-main} reports the terminal gaps. The learned policy attains the smallest gap in all six configurations. Against the envelope it improves the geometric-mean terminal gap by $23.6\times$ (range $4.5\times$ to $295\times$) and wins $40$ of the $48$ paired comparisons.

\section{Experiment Details for LP}
\label{app:pdlp_details}

This appendix collects the details deferred from \cref{sec:pdlp}: the PDHG backbone used in our implementation and its associated runtime state, the default FirstOrderLP restart logic, the feature design and architecture of the learned monitor, the synthetic LP generation and training setup, and the precise definition of each curve in \cref{fig:pdlp-main}.

\subsection{Problem Generation}
\label{app:lp_generation_and_alignment}

Our best small-scale result is obtained on a synthetic dense LP family with $(m,n)=(6,12)$.  Each instance is a standard-form primal LP
\begin{equation}
    \min_{x\in\RR^n} c^\top x
    \quad\text{s.t.}\quad
    Ax=b,\quad x\ge 0,
    \label{eq:primal-lp}
\end{equation}
with dual
\begin{equation}
    \max_{y \in \RR^m,\, s \in \RR^n} -b^\top y
    \quad \text{s.t.} \quad
    c+A^\top y=s,\quad s \ge 0.
    \label{eq:app-lp-dual}
\end{equation}
Then $s_{\rm d}(y^\star)=c+A^\top y^\star=s^\star$,
so $(x^\star,y^\star,s^\star)$ satisfies primal feasibility,
dual feasibility, and complementarity exactly.
A problem family is defined by the dimensions $(m,n)$.  We first sample a matrix $\widetilde A \in \RR^{m \times n}$ entrywise from $\cN(0,1)$ and normalize by its spectral norm:
\begin{equation}
    A = \widetilde A / \sigma_{\max}(\widetilde A).
\end{equation}
We then construct a certified primal--dual optimal solution $(x^\star,y^\star,s^\star)$ and back-solve for $(b,c)$.  Specifically, we sample an active support set $S \subseteq \{1,\dots,n\}$ of size
\begin{equation}
    |S| = \min\bigl(n,\max(m+2,\lfloor n/3 \rfloor)\bigr),
\end{equation}
set $x^\star_i \sim \mathrm{Unif}[0.25,1.25]$ for $i \in S$ and $x^\star_i=0$ otherwise, sample $y^\star \sim \cN(0,0.6^2 I_m)$, set $s^\star_i=0$ on $S$, and sample $s^\star_i \sim \mathrm{Unif}[0.25,1.25]$ on the inactive coordinates.  We then define
\begin{equation}
    b = A x^\star,
    \qquad
    c = s^\star - A^\top y^\star.
\end{equation}
By construction, $(x^\star,y^\star,s^\star)$ satisfies primal feasibility, dual feasibility, and complementarity exactly, so each generated instance comes with a certified KKT point.

For each LP instance, the initial state is
\begin{equation}
    z_0 = (x_0,y_0),
    \qquad
    x_0 \sim \mathrm{Unif}([-2.5,2.5]^n),
    \qquad
    y_0 \sim \mathrm{Unif}([-2.5,2.5]^m),
\end{equation}
with independent coordinates.

\subsection{PDHG Backbone and Restart Score}

We describe the primal--dual algorithm for LP as implemented in our codebase, following the FirstOrderLP(FOLP) algorithm from~\citet{applegate2023faster}.

\paragraph{Scaled state and PDHG update.}
The implementation first applies Ruiz scaling to the LP data: with diagonal row and column rescalings $D_r,D_c$, the scaled data are $\tilde A=D_r A D_c$, $\tilde b=D_r b$, $\tilde c=D_c c$, and the iterates $(\tilde x_k,\tilde y_k)$ live in the scaled coordinates $\tilde x=D_c^{-1}x$, $\tilde y=D_r^{-1}y$. The runtime maintains a scalar stepsize $s_k$ and primal weight $\omega_k$, with primal/dual stepsizes $\tau_k=s_k/\omega_k$ and $\sigma_k=s_k\omega_k$. The PDHG half-step is
\begin{equation}
\begin{aligned}
    \tilde x_{k+\frac12} &= \Pi_{\RR_+^n}\!\bigl(\tilde x_k-\tau_k(\tilde c+\tilde A^\top \tilde y_k)\bigr), \\
    \bar x_{k+\frac12}   &= 2\tilde x_{k+\frac12}-\tilde x_k, \\
    \tilde y_{k+\frac12} &= \tilde y_k+\sigma_k(\tilde A\,\bar x_{k+\frac12}-\tilde b).
\end{aligned}
    \label{eq:app-pdhg-half-step}
\end{equation}
An adaptive movement test compares a quadratic primal--dual displacement $m_k=\tfrac12\omega_k\nrm{\Delta x_k}^2+\tfrac12\omega_k^{-1}\nrm{\Delta y_k}^2$ to a cross-interaction term $q_k=\abs{\langle \Delta x_k,\tilde A^\top(\tilde y_{k+\frac12}-\tilde y_k)\rangle}$ and accepts the half-step iff $m_k\le \varepsilon$ or $s_k\le m_k/q_k$; otherwise $(\tilde x_{k+1},\tilde y_{k+1})$ is set to $(\tilde x_k,\tilde y_k)$. The stepsize is updated multiplicatively at every iteration toward the local stability limit. Throughout, the runtime maintains weighted running averages
\begin{equation}
    \tilde x_k^{\rm avg}=\frac{\sum_{t=1}^{k} s_t\tilde x_t}{\sum_{t=1}^k s_t},
    \qquad
    \tilde y_k^{\rm avg}=\frac{\sum_{t=1}^{k} s_t\tilde y_t}{\sum_{t=1}^k s_t},
    \label{eq:app-running-avg}
\end{equation}
which are used at restart checks and as the post-restart anchor.

\paragraph{Stepsize and primal-weight initialization.}
With the scaled data $(\tilde A,\tilde b,\tilde c)$, the runtime is initialized by
\begin{equation}
    s_0=\frac{1}{\max(\nrm{\tilde A}_\infty,1)},
    \qquad
    \omega_0=
    \begin{cases}
        \pi_{\rm imp}\cdot \frac{\nrm{\tilde c}_2}{\nrm{\tilde b}_2}, & \text{if both norms are nonzero}, \\
        \pi_{\rm imp}, & \text{otherwise},
    \end{cases}
    \label{eq:app-pdlp-init}
\end{equation}
where $\nrm{\tilde A}_\infty$ is the maximum absolute row sum of $\tilde A$ and $\pi_{\rm imp}$ is the \texttt{primal\_importance} hyperparameter. Define the displacement and cross-interaction
\begin{equation}
    m_k=\tfrac12\omega_k\nrm{\Delta\tilde x_k}_2^2+\tfrac12\omega_k^{-1}\nrm{\Delta\tilde y_k}_2^2,
    \qquad
    q_k=\abs{\langle \Delta\tilde x_k,\,\tilde A^\top(\tilde y_{k+\frac12}-\tilde y_k)\rangle},
\end{equation}
with per-step stability limit $s_k^{\max}=m_k/q_k$ when $q_k>\varepsilon$ and $s_k^{\max}=\infty$ otherwise. The stepsize is updated by
\begin{equation}
    s_{k+1}
    =\min\!\Bigl\{\bigl(1-(k+1)^{-a}\bigr)\,s_k^{\max},\,\bigl(1+(k+1)^{-b}\bigr)\,s_k\Bigr\},
    \label{eq:app-pdlp-stepsize-update}
\end{equation}
with defaults $a=0.3$, $b=0.6$.

\paragraph{KKT residual.}
For a state $z=(x,y)$ in the unscaled coordinates, define the normalized residual blocks
\begin{align}
    r_{\rm peq}(z)&=\tfrac{1}{\sqrt{m}}(Ax-b), &
    r_{\rm pineq}(z)&=\tfrac{1}{\sqrt{n}}[-x]_+, \nonumber \\
    r_{\rm dineq}(z)&=\tfrac{1}{\sqrt{n}}[-c-A^\top y]_+, &
    r_{\rm comp}(z)&=\tfrac{1}{\sqrt{n}}\,x\odot (c+A^\top y),
    \label{eq:app-residual-blocks}
\end{align}
where $[u]_+=\max(u,0)$ is applied elementwise. The KKT residual used as our evaluation and training metric is
\begin{equation}
    \cR(z)=\sqrt{\nrm{r_{\rm peq}(z)}_2^2+\nrm{r_{\rm pineq}(z)}_2^2+\nrm{r_{\rm dineq}(z)}_2^2+\nrm{r_{\rm comp}(z)}_2^2}.
    \label{eq:app-kkt-residual}
\end{equation}

\paragraph{Duality gap and proxy.}
The adaptive restart scheme of \citet{applegate2021practical} triggers a restart according to the \emph{normalized duality gap}
\begin{equation}
    \rho_r(z)\;=\;\frac{1}{r}\,\max_{\tilde z\in B_r(z)\cap Z}\bigl\{\cL(x,\tilde y)-\cL(\tilde x,y)\bigr\},
    \label{eq:app-ndg-original}
\end{equation}
where $B_r(z)=\{\tilde z\in Z:\nrm{z-\tilde z}\le r\}$ and $\cL(x,y)=c^\top x+y^\top(Ax - b)$.
Computing $\rho_r$ requires solving a trust-region subproblem at each restart check.
For LP, substituting the Lagrangian into \eqref{eq:app-ndg-original} yields
\begin{equation}
    \rho_r(z)
    =\frac{1}{r}\max_{\tilde x\ge 0,\,\tilde y:\,\nrm{(\tilde x-x,\tilde y-y)}_2\le r}
    \bigl\{
    \underbrace{(c+A^\top y)^\top(x-\tilde x)}_{\text{primal term}}
    +\underbrace{(Ax-b)^\top(\tilde y-y)}_{\text{dual term}}
    \bigr\}.
    \label{eq:app-ndg-lp}
\end{equation}
Since $\cL(x,\tilde y)-\cL(\tilde x,y)$ is linear in $(\tilde x,\tilde y)$, the trust-region maximization in \eqref{eq:app-ndg-lp} reduces to aligning $(\tilde x-x,\tilde y-y)$ with the residual vector $(-c-A^\top y,\;Ax-b)$ subject to $\tilde x\ge 0$ and $\nrm{(\tilde x-x,\tilde y-y)}\le r$.  When the non-negativity constraint $\tilde x\ge 0$ is inactive, the maximizer simply points along the residual direction, giving
\begin{equation}
    \rho_r(z)
    \;=\;
    \sqrt{\nrm{\bar s}_2^2+\nrm{Ax-b}_2^2},
    \label{eq:app-ndg-approx}
\end{equation}
where $\bar s=c+A^\top y$ is the dual slack (we write $\bar s$ to avoid confusion with the stepsize $s_k$).  In this regime, $\rho_r$ is the $\ell_2$ norm of the combined residual vector.

Motivated by this observation, our implementation replaces the trust-region solve with the following closed-form proxy.  Let $(\tilde x^r,\tilde y^r)$ denote the last restart anchor. For any restart candidate $z=(x,y)$ and anchor $\tilde z=(\tilde x,\tilde y)$, define the scaled distance
\begin{equation}
    \nrm{z-\tilde z}_k
    :=
    \sqrt{
        \frac{
            \omega_k \nrm{x-\tilde x}_2^2
            +
            \omega_k^{-1}\nrm{y-\tilde y}_2^2
        }{s_k}
    },
    \label{eq:app-pdlp-dist}
\end{equation}
and the LP-specific restart proxy
\begin{equation}
 \begin{aligned}
    \hat{\rho}(z; \tilde z)
    &={}
    \abs{c^\top x+b^\top y}
    \\ &\quad+
    \max(\nrm{z-\tilde z}_k,1)\,
    \bigl(
        \nrm{r_{\rm peq}(z)}_2+\nrm{r_{\rm pineq}(z)}_2+\nrm{r_{\rm dineq}(z)}_2+\nrm{r_{\rm comp}(z)}_2
    \bigr).
 \end{aligned}
    \label{eq:app-folp-gap-proxy}
\end{equation}
At a primal--dual feasible pair, $c^\top x+b^\top y=x^\top\bar s\geq0$ is the primal--dual objective gap. At an infeasible pair, its absolute value is only an objective-mismatch statistic and need not bound the localized Lagrangian gap. The residual blocks separately measure feasibility and complementarity. The $\max(\nrm{z-\tilde z}_k,1)$ factor weights their contribution by the distance from the phase anchor, with a floor of one; it is not derived from the $1/r$ normalization in \eqref{eq:app-ndg-original}.

We use $\widehat\rho$ as a computationally cheap proxy for the localized
normalized duality gap used in PDLP.  The proxy is designed to capture the same
qualitative residual information used by the restart rule, including primal
feasibility, dual feasibility, complementarity, and primal--dual objective
mismatch.  However, it does not solve the trust-region maximization problem that
defines the localized normalized duality gap in the original Julia
implementation.  Therefore, we do not claim that
$\widehat\rho(z;\tilde z)/\nrm{z-\tilde z}_k$ is equal or theoretically
equivalent to the exact PDLP gap measure.  We use it only as a low-cost restart
monitor in our Python implementation.  Once the residual components are
available, evaluating $\widehat\rho$ costs $O(n+m)$ and requires no inner solve,
which allows restart checks to be performed frequently.

\paragraph{Default FirstOrderLP restart rule.}
At a restart checkpoint, let $z^r=(\tilde x^r,\tilde y^r)$ denote the current restart anchor. The default \textsc{FOLP} logic first compares the current iterate and the averaged iterate through
\begin{equation}
    \phi_{\mathrm{cur}}
    =
    \frac{\hat\rho(z_k;\,z^r)}{\nrm{z_k-z^r}_k^2},
    \qquad
    \phi_{\mathrm{avg}}
    =
    \frac{\hat\rho(z_k^{\rm avg};\,z^r)}{\nrm{z_k^{\rm avg}-z^r}_k^2},
    \label{eq:app-folp-phi}
\end{equation}
where $z_k=(\tilde x_k,\tilde y_k)$ and $z_k^{\rm avg}=(\tilde x_k^{\rm avg},\tilde y_k^{\rm avg})$.
The restart candidate is whichever of the two has the smaller $\phi$ value.
If $z^{\mathrm{cand}}$ is the chosen candidate, define
\begin{equation}
    \rho_k
    =
    \frac{
        \hat\rho(z^{\mathrm{cand}};\,z^r) \,/\, \nrm{z^{\mathrm{cand}}-z^r}_k
    }{
        \hat\rho_{\mathrm{last}} \,/\, \nrm{\cdot}_{\mathrm{last}}
    },
    \label{eq:app-folp-rho}
\end{equation}
where $(\hat\rho_{\mathrm{last}}, \nrm{\cdot}_{\mathrm{last}})$ denote the gap proxy and distance of the previous restart segment. The default \textsc{FOLP} implementation restarts if either the artificial safeguard
\begin{equation}
    \ell_k \ge 0.5\,k,
    \label{eq:app-folp-artificial}
\end{equation}
where $\ell_k$ is the current phase length, or the normalized reduction test
\begin{equation}
    \rho_k < 0.9
    \quad\text{and}\quad
    \bigl(
        \rho_k < 0.1
        \;\text{or}\;
        \rho_k > \rho_{k-1}^{\mathrm{trial}}
    \bigr).
    \label{eq:app-folp-rho-test}
\end{equation}
After restart, the iterate is reset to the chosen candidate, the averaging buffers \eqref{eq:app-running-avg} are cleared, and the primal weight is updated from the primal/dual movement ratio with smoothing coefficient $0.5$.

\subsection{Learning Restart Policy for LP}

\paragraph{Network architecture.}
The policy is a small MLP $V(z;\theta)=\MLP(\phi(z);\theta)$ with 4 layers (embedding dimensions $10$, $8$, $7$, $4$) and ReLU activations. The feature vector $\phi(\xbar_k,\xbar_{N_i},k)\in\RR^{10}$ has entries being the logarithm of: the duality gap $\abs{c^\top\bar x_k + b^\top\bar y_k}$, the residual sum $\nrm{r_{\rm peq}}+\nrm{r_{\rm pineq}}+\nrm{r_{\rm dineq}}+\nrm{r_{\rm comp}}$, the primal/dual travel distances $d_{x,k}=\nrm{\tilde x_k^{\rm avg}-\tilde x_{N_i}}$ and $d_{y,k}=\nrm{\tilde y_k^{\rm avg}-\tilde y_{N_i}}$, the runtime stepsize $s_k$ and primal weight $\omega_k$, summary statistics $\widehat G_{k-1},\widehat D_{k-1},\widehat \rho_{k-1}$ from the previous phase, and the relative phase length $\varphi_k=(k-N_i)/\max(k,1)$. The first output coordinate is the restart logit; the remaining three outputs predict log-scale phase-summary statistics ($\log \widehat G$, $\log \widehat D$, $\log \widehat \rho$) used to update the monitor state after each restart. The total number of trainable parameters is $183$.

\paragraph{Training objective.}
The training objective is the weighted-path KKT residual
\begin{equation}
    J(\theta)
    =\EE_{P,z_0}\!\Biggl[\sum_{k=1}^{N} w_k\,\cR\bigl(\Xbar(k;\theta)\bigr)\Biggr],
    \qquad w_k=\frac{2k}{N(N+1)},
    \label{eq:app-pdlp-loss}
\end{equation}
with $\cR$ as in \eqref{eq:app-kkt-residual} and $w_k\propto k$ up-weighting later iterates. Training proceeds in two stages. We first \emph{warm-start} the network on a dataset of feature--label pairs $\mathcal{D}=\{(\phi_i,y_i)\}_{i=1}^{M}$ with $M=880$ examples collected from the baselines; the warm-start is fitted by minimizing the binary cross-entropy with restart logit $-V(\phi;\theta)$. Starting from the warm-started parameters, we then \emph{fine-tune} the network using the surrogate gradients of \cref{def:disc_restart_sensitivity} on the loss \eqref{eq:app-pdlp-loss} for $6$ epochs with AdamW (learning rate $3\times 10^{-4}$ and weight decay $10^{-4}$). The selected checkpoint is epoch~$5$, chosen by validation performance.

\paragraph{Baseline set.}
We compare against three FirstOrderLP baselines and their pointwise best envelope:
\begin{enumerate}[leftmargin=1.5em,itemsep=0.15em]
    \item \texttt{FOLP-default}: the default FirstOrderLP reference solver;
    \item \texttt{FOLP-freq200}: the same FirstOrderLP solver but with restart checks every $200$ iterations;
    \item \texttt{FOLP-no-restart}: the same FirstOrderLP solver with restart checks effectively disabled.
    \item \texttt{Best-baseline-env}: the pointwise-best KKT residual achieved by the above three baselines at each iteration, serving as a conservative comparison envelope.
\end{enumerate}

\begin{figure}[t]
    \centering
    \includegraphics[width=\linewidth]{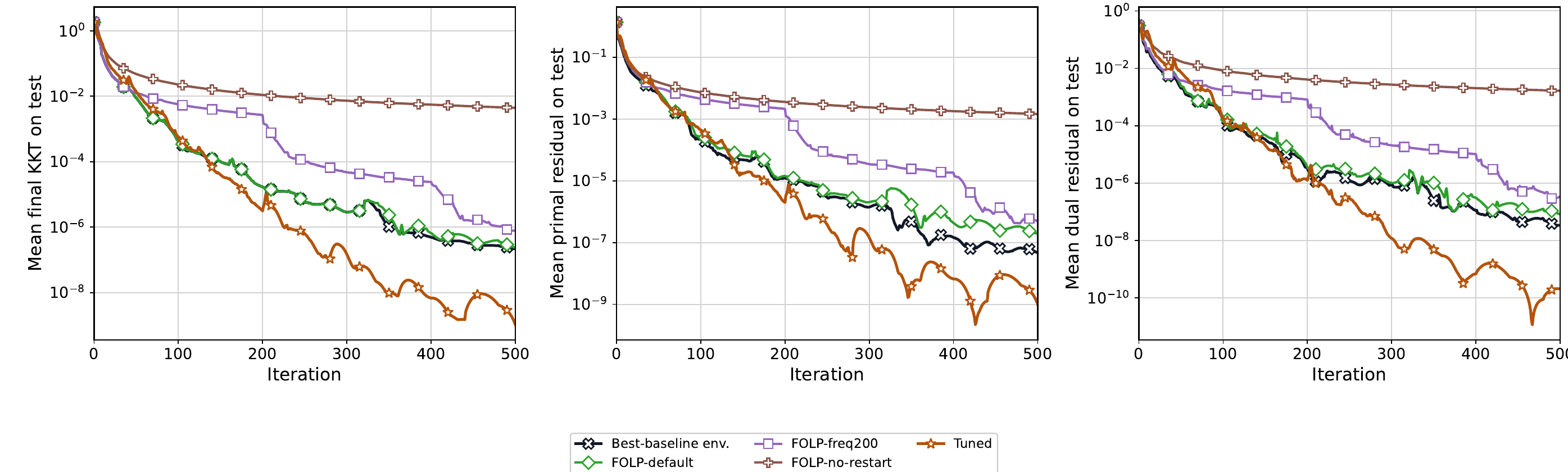}
    \vspace{-5pt}
    \caption{KKT residual (\emph{left}), primal residual (\emph{middle}), and dual residual (\emph{right}) vs.\ iteration for the learned restart policy and FirstOrderLP~\citep{applegate2023faster} baselines on synthetic LPs (dimension $n=18$, $m=6$). }
    \label{fig:pdlp-transfer}
\end{figure}

\paragraph{Out-of-distribution performance.}
To test generalization beyond the training scale, we evaluate the learned policy on synthetic LP instances with $n=18$ variables and $m=6$ constraints---$50\%$ larger than the training instances---without any additional fine-tuning. As shown in \cref{fig:pdlp-transfer}, the learned restart policy continues to match or improve upon the best-baseline envelope on these out-of-distribution instances, indicating that the policy captures transferable phase-level restart structure rather than overfitting to the training problem size.

\subsection{Netlib Benchmark: Policy Class, Training, and Evaluation}
\label{app:netlib}

This appendix details the Netlib experiment reported in \cref{sec:pdlp}. It differs from the synthetic-LP study above in three respects: the instances are real benchmark LPs, the policy class is an affine score rather than an MLP with monitor state, and the policy is trained on a small, a-priori-chosen subset of the benchmark and evaluated on the rest.

\paragraph{Policy class.}
Every $\texttt{check\_every}$ iterations the solver asks whether to restart, i.e.\ whether to take the running average as the new anchor. The learned score is affine in standardized features,
\begin{equation}
    V(\phi_k;\theta)=w^\top\frac{\phi_k-\mu}{\sigma}+b,
    \qquad\text{restart}\iff V(\phi_k;\theta)\le 0,
    \label{eq:app-netlib-score}
\end{equation}
with $\phi_k\in\RR^{15}$ the phase features listed below and $\theta=(w,b)\in\RR^{16}$ the only trainable parameters.

The reason for this restricted class is that it \emph{contains the handcrafted rule as a special case}: taking $w$ supported on the coordinate $\log\rho_k$ and $b=\mu_{14}-\log 0.85$ recovers the PDLP sufficient-decrease test $\rho_k\le 0.85$ exactly. Training can therefore be initialized at the baseline rule and can be read as improving a known rule inside a class that contains it, rather than fitting an unconstrained function that might be worse. With $16$ parameters and $7$ training instances the capacity is also small enough that overfitting is implausible, and the fitted weights are directly interpretable. We note explicitly that this is a simplification relative to the architecture of \cref{app:pdlp_details}; we adopt it because it suffices on these features, not because the MLP fails.

\paragraph{Features.}
The feature map $\phi_k\in\RR^{15}$ collects, all on a logarithmic scale: the duality gap and residual sum at the current iterate; the KKT residuals $R_{\rm cur}$ and $R_{\rm avg}$ of the current and averaged iterates; the squared primal and dual movements $\nrm{\Delta x}^2$, $\nrm{\Delta y}^2$ within the current phase; the stepsize $s$ and primal weight $\omega$; the candidate's normalized duality gap $\eta$, the value $\eta_{\rm last}$ from the previous phase, and the previous trial ratio $\rho_{\rm trial}$; the phase length $\ell$ and the relative phase length $\ell/k$; the per-iteration decrease rate of $\log\eta$; and finally the test statistic of the handcrafted rule, $\log\rho_k$ with $\rho_k=\eta/\eta_{\rm last}$, which is the coordinate that makes \eqref{eq:app-netlib-score} contain the PDLP rule.

\paragraph{Training instances.}
Of the $86$ Netlib instances that our parser handles, we take the $7$ with the fewest nonzeros in the MPS file: \texttt{afiro}, \texttt{sc50b}, \texttt{sc50a}, \texttt{kb2}, \texttt{sc105}, \texttt{adlittle}, \texttt{scagr7}. The selection rule is fixed in advance and uses no evaluation outcome. These seven instances are removed from every statistic we report. No mid- or large-tier instance is used for training, so both of those tiers are size extrapolation.

\paragraph{Randomized initial points.}
Rather than training from one fixed starting point, each rollout draws a fresh initialization,
\begin{equation}
    x_0\sim\mathrm{Unif}[0,x_{\rm hi})^n,\quad
    y_0\sim\cN(0,\sigma_y^2)^m,\quad
    x_{\rm hi}\sim\mathrm{LogUnif}[0.5,8],\quad
    \sigma_y\sim\mathrm{Unif}[0,2],
    \label{eq:app-netlib-init}
\end{equation}
so that the magnitudes themselves vary across rollouts and the policy sees a wide family of trajectories instead of a single one.

\paragraph{Objective and gradients.}
We minimize the same weighted-path KKT residual as in \eqref{eq:app-pdlp-loss}, in logarithmic form so that instances of different difficulty contribute on a common scale,
\begin{equation}
    \cL(\theta)=\sum_{k=1}^{N}w_k\log R_k,\qquad w_k=\frac{2k}{N(N+1)},
    \label{eq:app-netlib-loss}
\end{equation}
with the linearly increasing weights biasing the optimization towards terminal accuracy. Restart times are integer-valued and hence not differentiable in $\theta$; gradients are obtained with the surrogate of \cref{def:disc_restart_sensitivity}.

\paragraph{Optimization.}
The parameters are initialized at the PDLP rule as described above. We train for $40$ epochs with AdamW (learning rate $3\times10^{-4}$, weight decay $10^{-4}$, gradient-norm clipping at $1.0$). Checkpoints are selected on a fixed panel of $7$ instances $\times$ $3$ held-out initial points (seeds $2001$--$2003$) that never enter any gradient.

\paragraph{Evaluation protocol.}
Iteration budgets are $N=20000$, $25000$ and $50000$ for the small, mid and large tier, with $\texttt{check\_every}=1$. Each instance is run from $4$ independent random initial points (seeds $0$--$3$); a cell of \cref{tab:netlib-main} is the median across seeds within each instance, then the median across instances in the tier. To avoid letting non-converged runs dominate the medians, an instance is dropped from the comparison---for \emph{all} methods simultaneously---whenever any method ends above a KKT residual of $10^{5}$, leaving $n=32$, $24$ and $10$ instances. The baselines are the PDLP default rule, periodic restarting every $200$ iterations, and no restarting.

\end{document}

%% file: preamble.tex
\usepackage[utf8]{inputenc}
\usepackage[T1]{fontenc}
\usepackage{graphicx}
\usepackage[hypertexnames=false]{hyperref}
\usepackage{booktabs}
\usepackage{amsfonts}
\usepackage{microtype}
\usepackage{multirow}
\usepackage{enumitem}
\usepackage{breakcites}

\usepackage{algorithm}
\usepackage{algorithmic}

\usepackage{amsmath}
\usepackage{amsthm}
\usepackage{bbm}
\usepackage{amssymb}

\usepackage[nameinlink,capitalize]{cleveref}

\usepackage{mathtools}

\theoremstyle{plain}
\newtheorem{theorem}{Theorem}[section]
\newtheorem{proposition}[theorem]{Proposition}

\theoremstyle{definition}
\newtheorem{definition}[theorem]{Definition}
\newtheorem{assumption}[theorem]{Assumption}
\theoremstyle{remark}

\renewcommand{\eqref}[1]{\texorpdfstring{\hyperref[#1]{(\ref*{#1})}}{(\ref*{#1})}}

\crefformat{equation}{#2Eq.\,(#1)#3}
\Crefformat{equation}{#2Eq.\,(#1)#3}
\Crefformat{figure}{#2Figure~#1#3}
\Crefformat{assumption}{#2Assumption~#1#3}
\Crefname{assumption}{Assumption}{Assumptions}
\Crefformat{figure}{#2Figure #1#3}
\Crefformat{assumption}{#2Assumption #1#3}

\usepackage{crossreftools}
\makeatletter
\g@addto@macro\appendix{%
  \crefalias{section}{appendixsection}%
  \crefalias{subsection}{appendixsubsection}%
  \crefalias{subsubsection}{appendixsubsubsection}%
}
\makeatother
\crefname{appendixsection}{Appendix}{Appendices}
\Crefname{appendixsection}{Appendix}{Appendices}
\crefname{appendixsubsection}{Appendix}{Appendices}
\Crefname{appendixsubsection}{Appendix}{Appendices}
\crefname{appendixsubsubsection}{Appendix}{Appendices}
\Crefname{appendixsubsubsection}{Appendix}{Appendices}

%% file: macros.tex
\DeclarePairedDelimiter{\abs}{\lvert}{\rvert}
\DeclarePairedDelimiter{\crl}{\{}{\}}
\DeclarePairedDelimiter{\prn}{(}{)}
\DeclarePairedDelimiter{\nrm}{\|}{\|}
\DeclarePairedDelimiter{\tri}{\langle}{\rangle}

\def\ddefloop#1{\ifx\ddefloop#1\else\ddef{#1}\expandafter\ddefloop\fi}
\def\ddef#1{\expandafter\def\csname c#1\endcsname{\ensuremath{\mathcal{#1}}}}
\ddefloop ABCDEFGHIJKLMNOPQRSTUVWXYZ\ddefloop
\def\RR{\ensuremath{\mathbb{R}}}
\def\EE{\ensuremath{\mathbb{E}}}

%% file: related_work.tex
\subsection{Related Work}
\label{sec:related_work}

\paragraph{Adaptive restarts.}
For accelerated gradient methods, \citet{o2015adaptive} introduced two simple adaptive restart criteria and showed that they recover a linear rate on strongly convex problems without knowledge of the strong-convexity constant. Convergence guarantees for these and related rules have then been refined in~\citep{giselsson2014monotonicity,fercoq2019adaptive,fercoq2022quadratic}. For primal--dual methods on linear programming, adaptive restart policies are the central mechanism behind the practical performance of PDLP and its variants~\citep{applegate2021practical,applegate2023faster,lu2025geometry}, with worst-case and sharpness-based analyses developed in~\citep{hinder2024worst,applegate2023faster}. All of these rules are derived analytically and are governed by a small number of hand-tuned parameters.

\paragraph{Learning to optimize.}
L2O is a paradigm in which optimization algorithms are designed by training a parameterized model on a distribution of problem instances; see~\citep{chen2022learning,chen2024learning} for surveys. One line of work parameterizes either the per-step update directly by a neural network~\citep{l2o-dm,rnnprop} or the algorithm's coefficients by a small parametric family motivated by an ODE limit~\citep{xie2024ode}. Another line of work studies when L2O can come with convergence and generalization guarantees~\citep{liu2023towards,yang2023learninggeneralizeprovablylearning,yang2023ml2o}. The dominant training objective in both lines is to minimize the loss reached after a fixed iteration budget in the absence of \emph{restarts}, which is differentiable in the algorithm parameters and admits standard gradient-based meta-learning.

\paragraph{ODE perspectives on optimization.}
The gradient flow perspective provides both theoretical justification and a practical design guide for gradient-based algorithms~\citep[e.g.,][]{helmke2012optimization,wibisono2016variational} by relating discrete iterations to their continuous-time limit. Lyapunov analyses of these continuous-time models translate directly into convergence guarantees for the corresponding discrete iterations~\citep{laborde2020lyapunov,moucer2023systematic}, and a canonical instance is the ODE limit of Nesterov's accelerated gradient methods~\citep{su2016differential,shi2022understanding}. Building on this perspective, recent work~\citep{xie2024ode,xie2025accelerating} formulates a differentiable notion of stopping time for iterative algorithms and uses it to optimize convergence speed against a target accuracy. The underlying machinery---differentiating through state-dependent events of an ODE solve---is closely related to neural event functions for ODEs~\citep{chen2020learning} and, more broadly, to implicit differentiation through optimization procedures~\citep{blondel2022efficient,bertrand2022implicit,bolte2022nonsmoothimplicitdifferentiationmachine,xu2023revisiting}.

%% file: experiment.tex
\section{Experiments}
\label{sec:experiments}

We conduct experiments on two problem classes, QP with the NAG backbone and LP with the PDHG backbone of PDLP. All problem constructions, feature maps, and training hyperparameters are deferred to \cref{app:QP,app:logreg,app:ct,app:pdlp_details}.

\subsection{Learning to Restart for QP}
\label{sec:qp-discrete}

Throughout this subsection, the optimizer is NAG and we compare the learned restart policy against two standard handcrafted adaptive restart criteria from~\citet{o2015adaptive}: \textbf{Function restart} (restart iff $f(x_{k+1})>f(x_k)$) and \textbf{Gradient restart} (restart iff $\nabla f(y_k)^\top(x_{k+1}-x_k)>0$), together with \textbf{No restart}.

\begin{figure}[t]
    \centering
    \includegraphics[width=\linewidth,height=1.3in]{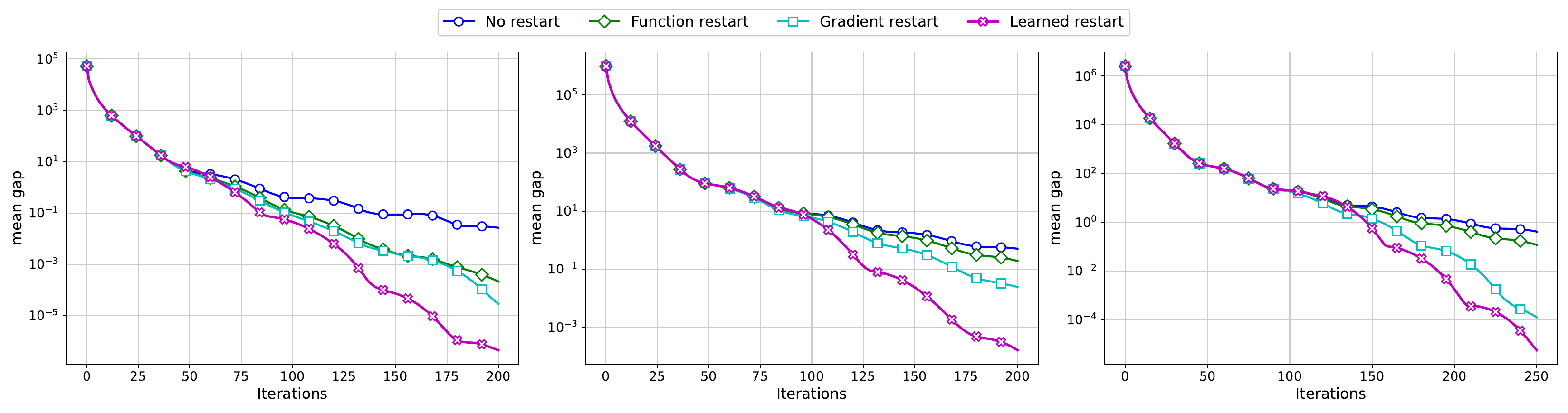}
    \vspace{-14pt}
    \caption{Learning a restart policy for NAG on the QP family~\eqref{eq:QP}: mean objective gap $f(x_k)-\min f$ along the trajectory for a policy trained on $d=20$ and evaluated on $d=20$ (\emph{left}), $400$ (\emph{middle}), and $1000$ (\emph{right}).}
    \label{fig:qp-id-main}
\end{figure}

\paragraph{Synthetic quadratics.}
\label{sec:qp-synthetic}
We evaluate the proposed differentiable restart mechanism by learning a restart policy for NAG on a family of diagonal QPs~\eqref{eq:QP}. The restart policy is a small two-layer MLP ($4$-$32$-$1$) taking four phase-level features as input. Training uses $20$-dimensional problems with a two-stage procedure: warm-start from function-restart teacher labels, followed by end-to-end fine-tuning on a ratio objective. Full details are in \cref{app:qp-discrete}.
\Cref{fig:qp-id-main} reports the mean objective gap along the optimization trajectory, averaged over test problems. Results on unseen condition numbers are reported in \cref{app:qp-discrete}.

\paragraph{Zero-shot transfer to real data.}
\label{sec:logreg}\label{sec:ct}
We next ask whether a policy trained \emph{only on synthetic QPs} transfers to real problems without retraining. We train a $7$-$32$-$1$ MLP on synthetic quadratics with the same two-stage procedure and test it on two real problem families: $\ell_2$-regularized logistic regression on five LIBSVM datasets~\citep{libsvm} and ridge reconstruction from physically measured X-ray tomography data~\citep{walnutdata,cheesedata} in dimensions up to $d=16384$. Although both tasks are out of distribution, the learned policy still attains the smallest terminal gap in every configuration. \Cref{tab:ct-main} reports the tomography results; the logistic-regression results and all experimental details are given in \cref{app:logreg,app:ct}.

\begin{table}[t]
\centering\footnotesize\setlength{\tabcolsep}{6pt}
\caption{Terminal gap $f(x_N)-f^\star$ on real-measurement CT ridge reconstruction~\eqref{eq:ct} ($N=800$; geometric mean over $8$ initial points). Per dataset, the two rows are the regularization levels $\lambda_{\rm rel}=3\cdot10^{-4}$ and $10^{-4}$. Details in \cref{app:ct}.}
\label{tab:ct-main}
\begin{tabular}{lrrrr}
\toprule
Dataset & No restart & Function & Gradient & Learned\\
\midrule
\multirow{2}{*}{\texttt{walnut82}}
   & 6.90e-09 & 2.46e-09 & 5.17e-11 & \textbf{1.76e-13}\\
 & 1.72e-08 & 1.72e-08 & 1.72e-08 & \textbf{2.65e-10}\\
\addlinespace[2pt]
\multirow{2}{*}{\texttt{cheeseL45}}
   & 1.56e-07 & 9.75e-11 & 4.91e-11 & \textbf{1.97e-12}\\
 & 5.72e-07 & 5.81e-08 & 5.01e-08 & \textbf{8.95e-09}\\
\addlinespace[2pt]
\multirow{2}{*}{\texttt{cheeseF45}}
   & 2.00e-07 & 3.50e-11 & 3.23e-11 & \textbf{2.27e-12}\\
 & 7.81e-07 & 8.35e-08 & 7.92e-08 & \textbf{1.77e-08}\\
\bottomrule
\end{tabular}
\end{table}

\subsection{Learning to Restart for LP}
\label{sec:pdlp}

\begin{figure}[t]
    \centering
    \includegraphics[width=\linewidth,height=1.4in]{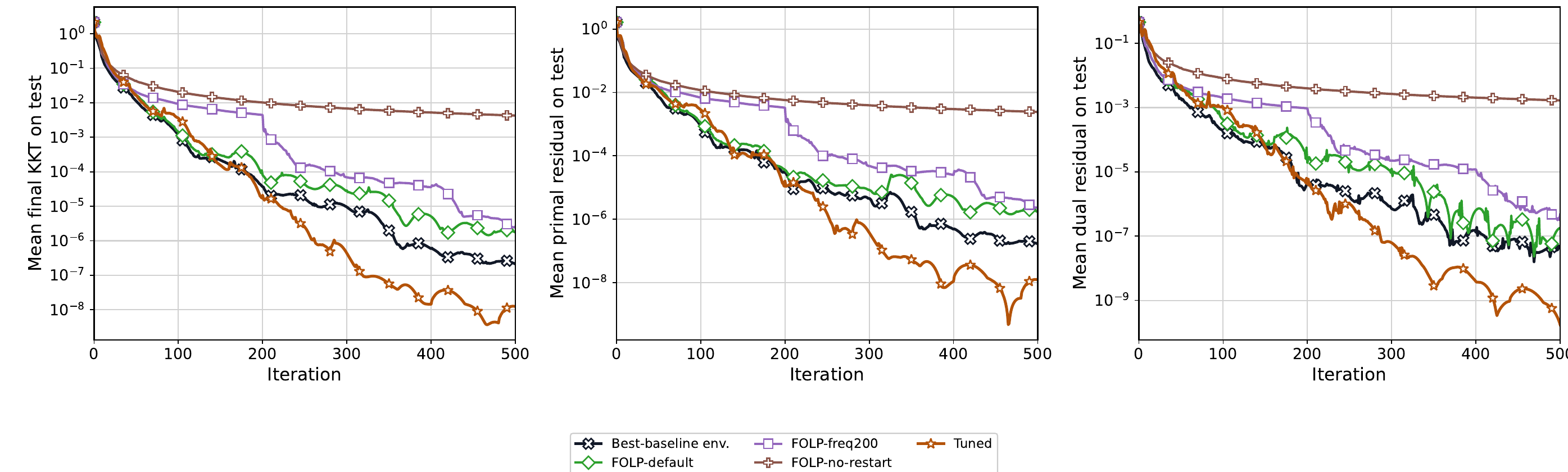}
    \vspace{-12pt}
    \caption{KKT residual (\emph{left}), primal residual (\emph{middle}), and dual residual (\emph{right}) vs.\ iteration for the learned restart policy and PDLP-style baselines on synthetic LPs (dimension $n=12$, $m=6$).}
    \label{fig:pdlp-main}
\end{figure}

We next apply our framework to primal--dual hybrid gradient (PDHG) methods on standard-form LPs, $\min_{x\in\RR^n} c^\top x$ s.t.\ $Ax=b$, $x\ge 0$, using a Python reimplementation of the PDHG backbone and adaptive restart scheme of PDLP~\citep{chambolle2011first,applegate2021practical}; the update and restart rule are given in \cref{app:pdlp_details}. Throughout, we compare against the PDLP restart rule, periodic restarting, and no restarting, together with their instance-wise best envelope. The training procedure is the same as for QP; all training details are in \cref{app:pdlp_details}.

\paragraph{Synthetic LPs.}
\label{sec:lp-synthetic}
We parametrize the restart score by a small MLP, $\Vbar_k=\MLP\bigl(\phi(\xbar_k,\xbar_{N_i},k);\theta\bigr)$, with $183$ trainable parameters.  \Cref{fig:pdlp-main} reports the KKT, primal, and dual residuals on held-out instances: the learned policy improves upon the best-baseline envelope in all three metrics, and the improvement carries over to larger unseen instances ($6\times 18$) without retraining (\cref{fig:pdlp-transfer}).

\paragraph{Netlib benchmark.}
\label{sec:netlib}
Finally, we train a restart policy directly on real instances from the Netlib LP benchmark. Here we deliberately use a \emph{minimal} policy class: the restart score is an affine function of $15$ log-scale phase features, with only $16$ trainable parameters. The policy is trained with the surrogate gradients on the $7$ smallest instances and evaluated on the remaining instances across three problem-size tiers (\cref{app:netlib}). \Cref{tab:netlib-main} reports the result. The learned policy is best in every tier and residual.

\begin{table}[!t]
\centering\footnotesize\setlength{\tabcolsep}{5pt}
\caption{Netlib LP benchmark: median final residual under a fixed iteration budget; $n$ is the number of retained instances per tier.}
\label{tab:netlib-main}
\begin{tabular}{llrrrrr}
\toprule
Tier & Residual & Learned & PDLP default & Periodic-200 & No restart & vs.\ default\\
\midrule
\multirow{3}{*}{\shortstack[l]{small\\$n{=}32$}}
 & KKT    & \textbf{2.29e-3} & 3.45e-3 & 4.62e-2 & 1.80e-1 & $1.51\times$\\
 & primal & \textbf{1.73e-5} & 1.85e-5 & 5.06e-3 & 1.48e-4 & $1.07\times$\\
 & dual   & \textbf{3.16e-6} & 4.47e-6 & 6.73e-6 & 6.14e-5 & $1.41\times$\\
\multirow{3}{*}{\shortstack[l]{mid\\$n{=}24$}}
 & KKT    & \textbf{4.83e-4} & 1.05e-3 & 2.37e-2 & 1.60e-2 & $2.16\times$\\
 & primal & \textbf{5.61e-5} & 9.88e-5 & 4.12e-3 & 2.12e-4 & $1.76\times$\\
 & dual   & \textbf{3.86e-6} & 9.91e-6 & 5.26e-5 & 6.87e-5 & $2.57\times$\\
\multirow{3}{*}{\shortstack[l]{large\\$n{=}10$}}
 & KKT    & \textbf{1.25e-3} & 1.43e-2 & 3.63e-2 & 2.26e-1 & $11.45\times$\\
 & primal & \textbf{8.71e-7} & 3.18e-6 & 6.87e-4 & 2.38e-5 & $3.65\times$\\
 & dual   & \textbf{4.85e-5} & 3.31e-4 & 8.71e-4 & 5.68e-5 & $6.82\times$\\
\bottomrule
\end{tabular}
\end{table}

%% file: revision/forward-implementation.tex
\begingroup
\subsection{Efficient Implementation by Forward Sensitivity Propagation}
\label{app:forward-implementation}
The surrogate recursions admit a single forward pass that propagates sensitivities alongside the optimizer state for update rule of the form $Z_{k+1}=\Phi_k(Z_k;\theta)$. This realizes the same quantities as \cref{app:auto-diff} without retaining a computation graph for the whole trajectory. We describe the propagation in the following.

Let $p=\dim\theta$ and let $Z_k\in\mathbb R^m$ contain the current optimizer state $X_k$, restart score $V_k$, and any auxiliary variables needed for a one-step update. Within phase $i$, write
\[
 Z_{k+1}=\Phi_k(Z_k;\theta),\qquad
 Z_0=\mathcal I(\xbar_i;\theta).
\]
Here $k$ is the local step index, held fixed when differentiating. The initialization map $\mathcal I$ sets the score and auxiliary variables, including any prescribed initial momentum.

Let $G_i=\disc{\xbar_i}\in\mathbb R^{D\times p}$ and initialize
\begin{equation}
 H_0=D_x\mathcal I(\xbar_i;\theta)G_i
             +D_\theta\mathcal I(\xbar_i;\theta).
 \label{eq:forward-initialization}
\end{equation}
Propagate the $m\times p$ sensitivity matrix by
\begin{equation}
 H_{k+1}=D_Z\Phi_k(Z_k;\theta)H_k+D_\theta\Phi_k(Z_k;\theta).
 \label{eq:forward-sensitivity-update}
\end{equation}
This is the ordinary chain rule at fixed local indices, with the incoming surrogate $G_i$ as its initial tangent. In particular, its state and score rows satisfy
\[
 H_k^{\sf X}=\partial_x\Fbarx(k;\xbar_i,\theta)G_i
                     +\partial_\theta\Fbarx(k;\xbar_i,\theta),\qquad
 H_k^{\sf V}=\partial_x\Fbarv(k;\xbar_i,\theta)G_i
                     +\partial_\theta\Fbarv(k;\xbar_i,\theta).
\]

At the restart step $k=S_i$, set
\begin{align}
 \Delta_i&=V_k-V_{k-1}<0,&
 d_i&=-H_k^{\sf V}/\Delta_i=\disc{S_i},
 \label{eq:forward-event-update}\\
 \xbar_{i+1}&=R(X_k),&
 G_{i+1}&=DR(X_k)H_k^{\sf X}
                  +\bigl(R(X_k)-R(X_{k-1})\bigr)d_i.
 \label{eq:forward-reset-update}
\end{align}
Substituting the expression for $H_k^{\sf X}$ and applying the chain rule to $R\circ\Fbarx$ shows that \eqref{eq:forward-reset-update} is exactly \eqref{eq:disc_dxi1_dp}. Reinitialize $Z_0,H_0$ with \eqref{eq:forward-initialization} for the next phase. Thus neither past phase snapshots nor past computation graphs are needed after this update.

\paragraph{Fixed global-time outputs and losses.}
Maintain the row vector $b_i=\disc{N_i}$, starting from $b_0=0$, and update it at a restart by $b_{i+1}=b_i+d_i$. At an interior step $n=N_i+k$, $k\geq1$, the global-trajectory surrogate is
\begin{equation}
 J_n:=\disc{\Xbar(n)}=H_k^{\sf X}-(X_k-X_{k-1})b_i.
 \label{eq:forward-global-output}
\end{equation}
This reproduces \eqref{eq:disc_dXn}; applying only \eqref{eq:forward-sensitivity-update} would omit the shift of earlier phase boundaries. The matrix $J_n$ is used for the output; For a differentiable loss $\ell(\Xbar(n),\theta)$ evaluated at an interior step $n$, its surrogate row gradient is
$D_X\ell\,J_n+D_\theta\ell$.

\paragraph{Runtime and memory.}
If $C_\Phi$ is the cost of one state update, $C_{\rm JVP}$ the cost of one such directional derivative, and $C_R$ the cost of a reset together with its sensitivity update, the total cost for $N$ iterations and $q$ restarts is
\[
 O\!\left(N(C_\Phi+pC_{\rm JVP})+qC_R\right).
\]
The additional sensitivity storage is $O(mp+p)$, with $O(m)$ storage for the current and preceding states and the temporary workspace for a single update. Thus, storage is independent of trajectory length and runtime is linear in the iteration count.
\endgroup

%% file: revision/restart-proofs.tex
\begingroup
\subsection{Proof of \cref{thm:restart_time_sensitivity}}
\label{app:restart_sensitivity_derivation}
\begin{proof}
Let $L_i^0=L_i(\theta_0)$ and $H_i(t,\theta)=\Fv(t;x_i(\theta),\theta)$. Starting from $D_\theta x_0=D_\theta T_0=0$, induction and the implicit function theorem give
\[
 \left.
 \begin{gathered}
 x_i,T_i\in C^1,\qquad H_i(L_i^0,\theta_0)=0,\\
 \partial_tH_i(L_i^0,\theta_0)<0
 \end{gathered}
 \right\}
 \Longrightarrow
 \begin{gathered}
 \ell_i\in C^1(U_i),\quad \ell_i(\theta_0)=L_i^0,\\
 H_i(\ell_i(\theta),\theta)=0.
 \end{gathered}
\]
For sufficiently small $r_i>0$ and a smaller neighborhood $U_i$,
\begin{align*}
 0<r_i<L_i^0,\qquad
 m_i&:=\min_{0\leq t\leq L_i^0-r_i}H_i(t,\theta_0)>0,\\
 H_i(t,\theta)&\geq m_i/2
      &&(0\leq t\leq L_i^0-r_i,\ \theta\in U_i),\\
 \partial_tH_i(t,\theta)&\leq-c_i<0
      &&(|t-L_i^0|\leq r_i,\ \theta\in U_i),\\
 |\ell_i(\theta)-L_i^0|&<r_i,\qquad
 H_i(t,\theta)>0\quad(0\leq t<\ell_i(\theta)).
\end{align*}
Hence $L_i=\ell_i$ on $U_i$, and
\begin{align*}
 0&=\partial_t\Fv\,D_\theta L_i
       +\partial_x\Fv\,D_\theta x_i+\partial_\theta\Fv,\\
 D_\theta L_i&=-\frac{\partial_x\Fv\,D_\theta x_i
                               +\partial_\theta\Fv}{\partial_t\Fv},\\
 D_\theta x_{i+1}&=\partial_t\Fr\,D_\theta L_i
                    +\partial_x\Fr\,D_\theta x_i+\partial_\theta\Fr,\\
 D_\theta T_{i+1}&=D_\theta T_i+D_\theta L_i.
\end{align*}
All derivatives of $F$ and $\Fr$ in this display are evaluated at $(L_i;x_i,\theta)$. For fixed $t$ strictly inside phase $i$, shrink the parameter neighborhood so that $T_i(\theta)<t<T_{i+1}(\theta)$. Then
\[
 D_\theta X(t)
 =-\partial_t\Fx(t-T_i;x_i,\theta)D_\theta T_i
   +\partial_x\Fx(t-T_i;x_i,\theta)D_\theta x_i
   +\partial_\theta\Fx(t-T_i;x_i,\theta).
\]
In the final phase, strict positivity of the score on $[0,T-T_q]$ and compactness give a positive margin. By continuity, this margin persists on a slightly longer interval and in a smaller parameter neighborhood, so no additional event enters the horizon. This also justifies the fixed-phase differentiation at $t=T$.
\end{proof}

\subsection{Monotone improvement}

\input{revision/monotone-improvement}
\input{revision/monotone-improvement-proof}

\subsection{Detailed Formulation and Proof of \cref{thm:disc_cont_close}}
\label{app:disc_cont_close_proof}
\input{revision/detailed-consistency}
\begin{proof}[Proof of \cref{thm:disc_cont_close-detailed}]
Empty maxima below are zero. Set $r_h=\epsilon_h+h$, $a_i=\partial_\theta x_i$, $\bar a_i=\disc{\xbar_i}$, $u_i=\|\xbar_i-x_i\|$, and $w_i=\|\bar a_i-a_i\|$. Constants are independent of $h$; for a family $F_h$, use the uniform margins stated above. The finite recursion \eqref{eq:dLi_dp}--\eqref{eq:dxi1_dp} gives
\[
 \max_{i<q}\|\partial_\theta L_i\|
 +\max_{i\leq q}\bigl(\|a_i\|+\|\partial_\theta T_i\|\bigr)\leq C,
 \qquad u_0=w_0=0.
\]
The forward-difference assumption also controls the backward differences used by the surrogate. For fixed $\delta>0$ and $h<\delta/2$, applying it at $t_{k-1}$ and using the time-Lipschitz bound on $[\delta/2,H]$ gives
\[
 \left\|\frac{\Fbar(k;x,\theta)-\Fbar(k-1;x,\theta)}h
             -\partial_tF(t_k;x,\theta)\right\|
 \leq C_{\delta/2}\epsilon_h+M_{\delta/2}h
 =O_\delta(r_h),\qquad \delta\leq t_k\leq H.
\]

\paragraph{First crossings.}
Write $v_i(t)=\Fv(t;x_i,\theta)$ and choose $0<\rho<L_i$ with
\[
 \min_{[0,L_i-\rho]}v_i=m_i>0,\qquad
 v_i'(t)\leq-\gamma/2\quad(|t-L_i|\leq\rho).
\]
By \eqref{eq:Fhat_consistency-detailed},
\begin{equation}
 |\Fbarv(k;\xbar_i,\theta)-v_i(kh)|
 \leq Cu_i+\epsilon_h=:b_i.
 \label{eq:revised-score-error}
\end{equation}
For $b_i<m_i$ and $2b_i/\gamma+h<\rho$,
\begin{align*}
 v_i(t)&\geq\tfrac\gamma2(L_i-t)
       &&(L_i-\rho\leq t\leq L_i),\\
 v_i(t)&\leq-\tfrac\gamma2(t-L_i)
       &&(L_i\leq t\leq L_i+\rho),\\
 kh<L_i-2b_i/\gamma&\ \Longrightarrow\ \Fbarv(k;\xbar_i,\theta)>0,\\
 k_i^+:=\left\lceil\frac{L_i+2b_i/\gamma}{h}\right\rceil
 &\ \Longrightarrow\ \Fbarv(k_i^+;\xbar_i,\theta)\leq0.
\end{align*}
Consequently,
\begin{equation}
 L_i-2b_i/\gamma\leq hS_i\leq hk_i^+\leq L_i+2b_i/\gamma+h,
 \qquad |hS_i-L_i|\leq C(u_i+r_h).
 \label{eq:revised-event-location}
\end{equation}
The reset and accumulated-time errors satisfy
\begin{align*}
 u_{i+1}
 &\leq C\bigl(\epsilon_h+u_i+|hS_i-L_i|\bigr)
 \leq C(u_i+r_h),\\
 \max_{i\leq q}u_i+\max_{i<q}|hS_i-L_i|&\leq Cr_h,\qquad
 |hN_i-T_i|\leq\sum_{j<i}|hS_j-L_j|\leq Cr_h.
\end{align*}
For the final-phase positive margin $m_q$ on $[0,T-T_q+\rho]$,
\[
 kh\leq T-hN_q
 \ \Longrightarrow\ kh\leq T-T_q+Cr_h,\qquad
 \Fbarv(k;\xbar_q,\theta)\geq m_q-Cu_q-\epsilon_h>0.
\]

\paragraph{Event sensitivities.}
Define
\begin{align*}
 d_i&=\partial_t\Fv(L_i;x_i,\theta),&
 \bar d_i&=\Delta_i/h,\\
 A_i&=\partial_\theta\Fv(L_i;x_i,\theta)
              +\partial_x\Fv(L_i;x_i,\theta)a_i,\\
 \bar A_i&=\partial_\theta\Fbarv(S_i;\xbar_i,\theta)
              +\partial_x\Fbarv(S_i;\xbar_i,\theta)\bar a_i.
\end{align*}
The consistency estimates at $hS_i\geq L_i/2>0$ give
\[
 |\bar d_i-d_i|\leq Cr_h,\quad |\bar d_i|\geq\gamma/2,\quad
 \|A_i\|\leq C,\quad \|\bar A_i-A_i\|\leq C(r_h+w_i).
\]
Thus
\begin{equation}
 \left\|h\disc{S_i}-\partial_\theta L_i\right\|
 \leq\frac{\|\bar A_i-A_i\|}{|\bar d_i|}
 +\frac{\|A_i\|\,|\bar d_i-d_i|}{|d_i\bar d_i|}
 \leq C(r_h+w_i).
 \label{eq:revised-event-sensitivity}
\end{equation}

\paragraph{Reset sensitivities.}
Let $y=\Fbarx(S_i;\xbar_i,\theta)$, $y^-=\Fbarx(S_i-1;\xbar_i,\theta)$, and $z=\Fx(L_i;x_i,\theta)$. Then
\[
 y-z=O(r_h),\qquad
 y-y^-=h\partial_t\Fx(L_i;x_i,\theta)+O(hr_h)=O(h),
\]
\begin{align}
 R(y)-R(y^-)
 &=\left[\int_0^1DR\bigl(y^-+s(y-y^-)\bigr)\,ds\right](y-y^-)
 \notag\\
 &=DR(y)(y-y^-)+O(h^2)
 =h\partial_t\Fr(L_i;x_i,\theta)+O(hr_h).
 \label{eq:revised-reset-difference}
\end{align}
With $B_i=(R(y)-R(y^-))/h$,
\begin{align*}
 B_i&=\partial_t\Fr(L_i;x_i,\theta)+O(r_h),\\
 \partial_{(x,\theta)}\Fbarr(S_i;\xbar_i,\theta)
 &=DR(y)\partial_{(x,\theta)}\Fbarx(S_i;\xbar_i,\theta)\\
 &=\partial_{(x,\theta)}\Fr(L_i;x_i,\theta)+O(r_h),\\
 \bar a_{i+1}
 &=B_i\,h\disc{S_i}
   +\partial_x\Fbarr(S_i;\xbar_i,\theta)\bar a_i
   +\partial_\theta\Fbarr(S_i;\xbar_i,\theta),\\
 w_{i+1}&\leq C\bigl(r_h+w_i+
                  \|h\disc{S_i}-\partial_\theta L_i\|\bigr)
 \leq C(r_h+w_i).
\end{align*}
Finite induction yields
\[
 \max_{i\leq q}w_i
 +\max_{i<q}\|h\disc{S_i}-\partial_\theta L_i\|\leq Cr_h,
\]
which proves \eqref{eq:sensitivity_close_step-detailed}--\eqref{eq:sensitivity_close_state-detailed}.

\paragraph{Matched global times.}
For $\operatorname{dist}(t_n,\{0,T_1,\ldots,T_q\})\geq\delta$ and $Cr_h<\delta/2$,
\begin{gather*}
 t_n\in(T_i,T_{i+1})\ \Longrightarrow\ N_i<n<N_{i+1},\\
 k=n-N_i,\qquad \ell=t_n-T_i,\qquad
 |kh-\ell|=|hN_i-T_i|\leq Cr_h,\qquad kh,\ell\geq\delta/2,\\
 \|\Xbar(n)-X(t_n)\|\leq\epsilon_h+C(u_i+|kh-\ell|)\leq C_\delta r_h,\\
 \|h\disc{N_i}-\partial_\theta T_i\|
 \leq\sum_{j<i}\|h\disc{S_j}-\partial_\theta L_j\|\leq Cr_h.
\end{gather*}
For the final phase, the same implications hold with its right endpoint beyond $T$. Define
\[
 \bar Q_i=\frac{\Fbarx(k;\xbar_i,\theta)-\Fbarx(k-1;\xbar_i,\theta)}h,\quad
 Q_i=\partial_t\Fx(\ell;x_i,\theta),\quad
 J_i=\partial_x\Fx(\ell;x_i,\theta).
\]
By \eqref{eq:Fhat_consistency-detailed}--\eqref{eq:Vhat_time_consistency-detailed},
\begin{gather*}
 \bar Q_i-Q_i=O_\delta(r_h),\qquad
 \partial_{(x,\theta)}\Fbarx(k;\xbar_i,\theta)
 -\partial_{(x,\theta)}\Fx(\ell;x_i,\theta)=O_\delta(r_h).
\end{gather*}
\begin{align*}
 \disc{\Xbar(n)}-\partial_\theta X(t_n)
 &=-\bar Q_i\bigl(h\disc{N_i}-\partial_\theta T_i\bigr)
   -(\bar Q_i-Q_i)\partial_\theta T_i
   \\
 &\quad+J_i(\bar a_i-a_i)+O_\delta(r_h)
 =O_\delta(r_h).
\end{align*}
\end{proof}

\endgroup

%% file: revision/monotone-improvement.tex
\begingroup
\begin{theorem}[Monotone improvement]
\label{thm:monotone-improvement}
Let $\Theta\subset\mathbb R^p$ be open and convex, and let $x_0$ be a fixed initial state. Suppose the restart hypotheses of \cref{thm:restart_time_sensitivity} hold throughout $\theta\in\Theta$. Assume \cref{ass:disc_consistency}(1)--(2) hold uniformly, and $\mathcal L$ is uniformly Lipschitz and smooth. Then the terminal loss $\ell(\theta)=\mathcal L(X(T;\theta))$ is $M_\ell$-smooth on $\Theta$ for some $M_\ell>0$.

As a corollary, suppose $\inf_\Theta\ell> -\infty$ and $0<\eta<1/M_\ell$, projected gradient descent (PGD) with the exact terminal-loss gradient satisfies
\begin{align}
 \theta_{k+1}&=\Pi_\Theta\bigl(\theta_k-\eta\nabla\ell(\theta_k)\bigr),
 &G_\eta(\theta_k)&=\frac{\theta_k-\theta_{k+1}}{\eta},
 \label{eq:monotone-pgd}\\
 \ell(\theta_{k+1})&\leq\ell(\theta_k)
 -\eta\left(1-\frac{M_\ell\eta}{2}\right)\|G_\eta(\theta_k)\|^2.
 \label{eq:monotone-descent}
\end{align}
Thus every nonzero update strictly improves the loss, and $\|G_\eta(\theta_k)\|\to0$. If projection is eventually inactive, then $\|\nabla\ell(\theta_k)\|\to0$ as well.
\end{theorem}
\endgroup

%% file: revision/monotone-improvement-proof.tex
\begingroup
\begin{proof}
\label{app:monotone-improvement-proof}
There are $q$ restarts in $[0,T]$. Let $A\geq1$ bound the first derivatives and their Lipschitz constants in \cref{ass:disc_consistency}(2). For functions on $\Theta$, write $\|\cdot\|_\infty$ and $\operatorname{Lip}(\cdot)$ for the supremum norm and Lipschitz constant. By \cref{app:restart_sensitivity_derivation}, $x_i,L_i,T_i\in C^1(\Theta)$.

\paragraph{Phase sensitivities.}
Set $P_i=D_\theta x_i$, $b_i=D_\theta L_i$, $c_i=\|P_i\|_\infty$, and $d_i=\|b_i\|_\infty$. Equations~\eqref{eq:dLi_dp}--\eqref{eq:dxi1_dp} give
\begin{equation}
 c_0=0,\qquad d_i\leq\frac A\gamma(1+c_i),\qquad
 c_{i+1}\leq A^2(1+c_i+d_i),\qquad
 \|D_\theta T_i\|_\infty\leq\sum_{j<i}d_j.
 \label{eq:monotone-phase-bounds}
\end{equation}
For $z_i(\theta)=(L_i(\theta),x_i(\theta),\theta)$ and $\theta_s=\theta'+s(\theta-\theta')$,
\[
 z_i(\theta)-z_i(\theta')
 =\int_0^1Dz_i(\theta_s)(\theta-\theta')\,ds,
 \qquad \operatorname{Lip}(z_i)\leq D_i:=1+c_i+d_i.
\]
Define
\[
 \alpha_i=\partial_\theta\Fv(z_i)+\partial_x\Fv(z_i)P_i,
 \quad \nu_i=\partial_t\Fv(z_i),\quad
 \lambda_i=\operatorname{Lip}(P_i),\quad
 \kappa_i=\operatorname{Lip}(b_i).
\]
Then
\begin{align*}
 \|\alpha_i\|_\infty&\leq A(1+c_i),&
 \operatorname{Lip}(\nu_i)&\leq AD_i,&\nu_i&\leq-\gamma,\\
 \operatorname{Lip}(\alpha_i)&\leq AD_i(1+c_i)+A\lambda_i,&&&\\
 b_i=-\alpha_i/\nu_i
 \quad\Longrightarrow\quad
 \kappa_i&\leq\frac{AD_i(1+c_i)+A\lambda_i}{\gamma}
             +\frac{A^2(1+c_i)D_i}{\gamma^2}.&&&
\end{align*}
Moreover,
\begin{align*}
 D\Fr(z)&=DR(\Fx(z))D\Fx(z),\qquad \|D\Fr\|_\infty\leq A^2,\\
 \|D\Fr(z)-D\Fr(z')\|
 &\leq A\|D\Fx(z)-D\Fx(z')\|
       +A^2\|\Fx(z)-\Fx(z')\|\\
 &\leq C_R\|z-z'\|,\qquad C_R=A^2+A^3,\\
 \lambda_{i+1}&\leq C_RD_i(1+c_i+d_i)+A^2(\lambda_i+\kappa_i),
 \qquad\lambda_0=0.
\end{align*}
Thus $c_i,d_i,\lambda_i,\kappa_i<\infty$ for all realized phases, and
$\operatorname{Lip}(D_\theta T_q)\leq\sum_{j<q}\kappa_j$.

\paragraph{Terminal loss.}
Put $Y(\theta)=\Fx(T-T_q;x_q,\theta)$ and
$J=1+c_q+\sum_{j<q}d_j$. From \eqref{eq:dXt_dp},
\begin{gather*}
 DY=-\partial_t\Fx\,D_\theta T_q+\partial_x\Fx\,P_q+\partial_\theta\Fx,
 \qquad (t,x,\theta)=(T-T_q,x_q,\theta),\\
 B:=AJ,\qquad C:=AJ^2+A\left(\lambda_q+\sum_{j<q}\kappa_j\right),
\end{gather*}
\begin{equation}
 \|DY\|_\infty\leq B,\qquad
 \|DY(\theta)-DY(\theta')\|\leq C\|\theta-\theta'\|.
 \label{eq:monotone-terminal-jacobian}
\end{equation}
Consequently,
\begin{align*}
 \|Y(\theta)-Y(\theta')\|
 &=\left\|\int_0^1 DY(\theta_s)(\theta-\theta')\,ds\right\|
 \leq B\|\theta-\theta'\|,\\
 \nabla\ell(\theta)&=DY(\theta)^\top\nabla\mathcal L(Y(\theta)),
 \qquad \|\nabla\mathcal L\|_\infty\leq L_{\mathcal L},\\
 \|\nabla\ell(\theta)-\nabla\ell(\theta')\|
 &\leq B M_{\mathcal L}\|Y(\theta)-Y(\theta')\|
       +L_{\mathcal L}\|DY(\theta)-DY(\theta')\|\\
 &\leq(M_{\mathcal L}B^2+L_{\mathcal L}C)\|\theta-\theta'\|.
\end{align*}
Take $M_\ell=\max\{1,M_{\mathcal L}B^2+L_{\mathcal L}C\}$.

\paragraph{Projected descent.}
Together with the smoothness bounds, the standard results for proximal gradient descent gives the desired inequalities; see \citet[Lemma~10.4]{beck2017first}.
\end{proof}
\endgroup

%% file: revision/detailed-consistency.tex
\begingroup
We first prove a local version of the consistency theorem that allows nonuniform regularity near phase starts, as needed for NAG. Under the uniform assumptions in the main text the proof also applies.
\begin{assumption}
\label{ass:disc_consistency-detailed}
Fix a finite horizon $T$ and a slightly larger local-time interval $[0,H]$, $H>T$. All bounds below hold on a neighborhood of the realized phase initializations and parameters.

(1) The event hypotheses of \cref{thm:restart_time_sensitivity} hold, and $\partial_t\Fv(L_i;x_i,\theta)\leq-\gamma<0$ for $0\leq i<q$.

(2) The first derivatives of $F$ and $R$ are bounded by $M$, and $DR$ is $M$-Lipschitz. For every fixed $\delta>0$, the first derivatives of $F$ are jointly $M_\delta$-Lipschitz on local times $[\delta,H]$.

(3) For $t_k=kh\in[0,H]$, the value and initialization/parameter derivatives satisfy
\begin{equation}
 \|\Fbar(k;x,\theta)-F(t_k;x,\theta)\|
 +\|\partial_{(x,\theta)}\Fbar(k;x,\theta)
       -\partial_{(x,\theta)}F(t_k;x,\theta)\|\leq\epsilon_h.
 \label{eq:Fhat_consistency-detailed}
\end{equation}
For every fixed $\delta>0$, the forward time differences satisfy
\begin{equation}
 \left\|\frac{\Fbar(k+1;x,\theta)-\Fbar(k;x,\theta)}h
             -\partial_t F(t_k;x,\theta)\right\|
 \leq C_\delta\epsilon_h,\qquad \delta\leq t_k<t_{k+1}\leq H.
 \label{eq:Vhat_time_consistency-detailed}
\end{equation}
Here $\epsilon_h\to0$, and all constants are independent of $h$.
\end{assumption}

\begin{theorem}[Closeness of surrogate gradients to continuous gradients]
\label{thm:disc_cont_close-detailed}
Suppose \cref{ass:disc_consistency-detailed} holds, $\xbar_0=x_0$ is independent of $\theta$, and $h$ is sufficiently small. The discrete and continuous trajectories then have the same $q$ restarts in $[0,T]$, and
\begin{align}
 \abs{hS_i-L_i}+\nrm{h\disc{S_i}-\partial_\theta L_i}
   &=O(\epsilon_h+h), &&0\leq i<q,
   \label{eq:sensitivity_close_step-detailed}\\
 \nrm{\xbar_{i+1}-x_{i+1}}+
 \nrm{\disc{\xbar_{i+1}}-\partial_\theta x_{i+1}}
   &=O(\epsilon_h+h), &&0\leq i<q.
   \label{eq:sensitivity_close_state-detailed}
\end{align}
For every fixed $\delta>0$,
\begin{equation}
 \nrm{\Xbar(n)-X(t_n)}+
 \nrm{\disc{\Xbar(n)}-\partial_\theta X(t_n)}
   =O_\delta(\epsilon_h+h)
 \label{eq:sensitivity_close_traj-detailed}
\end{equation}
uniformly over $t_n=nh\leq T$ with
$\operatorname{dist}(t_n,\{0,T_1,\ldots,T_q\})\geq\delta$.
Such sample times belong to corresponding phases for small $h$.
The value and sensitivity at the fixed initial point agree exactly.
\end{theorem}

The theorem also applies when the continuous reference is a family $F=F_h$, as in the high-resolution NAG model, provided all constants and neighborhoods above are uniform. In addition, all regularity and consistency bounds can be realxed.

In \cref{app:restart_examples} and \cref{app:nag_consistency}, we establish consistency with $\epsilon_h=O(h)$ for forward Euler and $\epsilon_h=O(h\log(1/h))$ for NAG. For NAG, the phase coordinate is the initial position with prescribed initial velocity, and the time-difference estimate is restricted to positive local times as above. Under the event hypotheses, the corresponding surrogate sensitivities converge at these rates.
\endgroup

%% file: revision/nag-consistency.tex
\begingroup
\subsection{Consistency for NAG and its ODE}
\label{app:nag_consistency}
We compare the NAG iteration with $h=\sqrt{s}$ to the $h$-dependent high-resolution ODE \eqref{eq:NAG-C-ODE}. All initialization derivatives are with respect to the phase-start \emph{position}; the initial velocity is prescribed as a function of that position.

Write $g=\nabla f$, $H=\nabla^2f$, $t_k=kh$, and $\tau=3h/2$. The discrete and continuous dynamics are
\begin{align}
 y_k&=x_k+\beta_khv_k,\quad
 v_{k+1}=\beta_kv_k-hg(y_k),\quad
 x_{k+1}=x_k+hv_{k+1},\quad \beta_k=\frac{k-1}{k+2},
 \label{eq:app-nagc-iter}\\
 \dot x&=v,\qquad
 \dot v=-\frac3{t+\tau}v-hH(x)v
             -\left(1+\frac{3h}{2(t+\tau)}\right)g(x).
 \label{eq:app-nagc-ode}
\end{align}
At a phase start $a$, initialize the discrete state at $(a,hb_d(a))$ and the continuous state at $(a,hb_c(a))$. Thus $b_d=0$ describes a discrete momentum reset, while $b_c=-g$ gives the initialization in \eqref{eq:NAG-C-ODE}.

\begin{proposition}[NAG consistency on the prescribed initialization manifold]
\label{prop:nagc-state-consistency}
Let $f\in C^3(\mathbb R^d)$ be convex and assume global smoothness: for some $M>0$,
\[
 \|H(x)\|+\|DH(x)\|\leq M,\qquad x\in\mathbb R^d.
\]
Thus the gradient and Hessian are globally Lipschitz. Assume $b_c,b_d\in C^2(\mathbb R^d;\mathbb R^d)$ have globally bounded first and second derivatives. Fix $T>0$ and a bounded open set $U$ of initial positions. For a function $W$ of $a$, define
\[
 \|W\|_*:=\sup_{a\in U}\bigl(\|W(a)\|+\|D_aW(a)\|\bigr),
 \qquad \eta_h:=h\bigl[1+\log(1+T/h)\bigr],
\]
using Euclidean and induced operator norms. Put $Z(t)=(x(t),v(t))$ and $\bar Z_k=(x_k,v_k)$. Then
\begin{equation}
 \|\bar Z_k-Z(t_k)\|_*\leq C\eta_h,
 \qquad 0\leq k\leq\lfloor T/h\rfloor.
 \label{eq:app-nagc-state-consistency}
\end{equation}
For every fixed $\delta>0$, uniformly in $a\in U$,
\begin{equation}
 \left\|\frac{\bar Z_{k+1}-\bar Z_k}{h}-\dot Z(t_k)\right\|
 \leq C_\delta\eta_h,
 \qquad \delta\leq t_k<t_{k+1}\leq T.
 \label{eq:app-nagc-time-derivative-consistency}
\end{equation}
The constants are independent of $h$ and $k$.
\end{proposition}

The global derivative bounds are imposed for simplicity. They can be relaxed to the corresponding bounds on bounded convex neighborhoods containing the initializations, continuous and discrete trajectories, and all intermediate evaluation points, provided these neighborhoods are independent of $h$, $a\in U$, and the policy parameters under consideration.

\begin{proof}
Constants may depend on $T$, $U$, the global derivative bounds, and $g(0),b_c(0),b_d(0)$, but not on $0<h\leq1$.

\paragraph{Continuous bounds.}
Set $u=v+hg(x)$ and $z=(x,u)$. This removes the Hessian term from the second equation:
\begin{equation}
 \dot x=u-hg(x),\qquad
 \dot u=-\frac3{t+\tau}u-\frac{t}{t+\tau}g(x),\qquad
 u(0)=h(b_c(a)+g(a)).
 \label{eq:nag-transformed-ode}
\end{equation}
The integrating-factor formula is
\begin{equation}
 u(t)=\frac{\tau^3u(0)}{(t+\tau)^3}
       -\int_0^t\frac{s(s+\tau)^2}{(t+\tau)^3}g(x(s))\,ds.
 \label{eq:nag-integrating-factor}
\end{equation}
The kernel is at most one, and global smoothness gives $\|g(x)\|\leq\|g(0)\|+M\|x\|$. This formula and the equation for $x$ therefore yield $\sup_{t\leq T}(\|x(t)\|+\|u(t)\|)\leq C$ by Gronwall's inequality. Similarly, \eqref{eq:app-nagc-iter} gives
\[
 \|\bar Z_{k+1}\|\leq(1+Ch)\|\bar Z_k\|+Ch,
 \qquad \sup_{k\leq T/h}\|\bar Z_k\|\leq C.
\]
Thus the continuous states, discrete states, and extrapolated points are uniformly bounded on the finite horizon, and so is $g$ at these points. Differentiating \eqref{eq:nag-integrating-factor} in $a$, using $D_ax(0)=I$, and applying Gronwall's inequality gives $\sup_{t\leq T}\|D_ax(t)\|\leq C$ and $\|D_au(t)\|\leq C(t+h)$. Together with the undifferentiated formula, this yields
\begin{equation}
 \|x(t)\|_*\leq C,\qquad
 \|u(t)\|_*+\|v(t)\|_*\leq C(t+h),\qquad
 \|\dot z(t)\|_*\leq C.
 \label{eq:nag-continuous-bounds}
\end{equation}
Indeed, the integral equations bound $\|D_au(t)\|$ by $Ch+C\int_0^t\|D_ax(s)\|\,ds$ and $\|D_ax(t)\|$ by $1+\int_0^t(\|D_au(s)\|+Ch\|D_ax(s)\|)\,ds$. Differentiating \eqref{eq:nag-transformed-ode} in time now gives
\begin{equation}
 \|\ddot z(t)\|_*\leq C\left(1+\frac1{t+h}\right).
 \label{eq:nag-variation-time-bound}
\end{equation}
Only $g,H,DH$ enter these bounds. Twice differentiating the integral equations introduces the bounded forcing $DH(x)[D_ax,D_ax]$; the same Gronwall argument gives $\|D_a^2x(t)\|\leq C$ and $\|D_a^2u(t)\|\leq C(t+h)$.

\paragraph{One-step stability and error.}
In the transformed coordinates, the discrete map $\Psi_k:(x,u)\mapsto(x^+,u^+)$ is
\[
 \begin{aligned}
 v&=u-hg(x),\qquad y=x+\beta_khv,\qquad x^+=y-h^2g(y),\\
 u^+&=\beta_ku-\beta_khg(x)+h\{g(x^+)-g(y)\}.
 \end{aligned}
\]
The original NAG map has derivative $\operatorname{diag}(I,\beta_kI)+O(h)$ and second derivative $O(h)$. The change of coordinates and its inverse have derivative $I+O(h)$ and second derivative $O(h)$. Since $|\beta_k|\leq1$, the chain rule gives
\begin{equation}
 \|D\Psi_k\|\leq1+Ch,\qquad \|D^2\Psi_k\|\leq Ch.
 \label{eq:nag-discrete-stability}
\end{equation}
Along $z(t_k)$, write $d_k=\Psi_k(z(t_k))-z(t_k)-h\dot z(t_k)$. Direct subtraction gives
\begin{equation}
 \begin{aligned}
 d_k^x&=h(\beta_k-1)v-h^2g(y),\\
 d_k^u&=\frac{3h^2\{u+(t_k+h)g(x)\}}{2(t_k+\tau)(t_k+2h)}
          +h\{g(x^+)-g(y)\}.
 \end{aligned}
 \label{eq:nag-map-defect}
\end{equation}
Here $\beta_k-1=-3h/(t_k+2h)$ and $x^+-y=-h^2g(y)$. The latter identity and the Lipschitz bound on $H$ imply $\|g(x^+)-g(y)\|_*\leq Ch^2$. Consequently, \eqref{eq:nag-continuous-bounds} yields $\|d_k\|_*\leq C(h^2+h^2/(t_k+h))$. Taylor's integral formula and \eqref{eq:nag-variation-time-bound} therefore give
\begin{equation}
 \begin{aligned}
 r_k&:=z(t_{k+1})-\Psi_k(z(t_k))
       =\int_{t_k}^{t_{k+1}}(t_{k+1}-s)\ddot z(s)\,ds-d_k,\\
 \|r_k\|_*&\leq C\left(h^2+\frac{h}{k+1}\right).
 \end{aligned}
 \label{eq:nag-local-defect}
\end{equation}

\paragraph{Accumulation.}
Let $\bar z_k=(x_k,v_k+hg(x_k))$ and $E_k=\|\bar z_k-z(t_k)\|_*$. The initial discrepancy gives $E_0\leq Ch$. By \eqref{eq:nag-discrete-stability}, the chain rule, and the bound on $D_az(t_k)$,
\[
 E_{k+1}\leq(1+Ch)E_k+C\left(h^2+\frac h{k+1}\right),\qquad
 E_k\leq e^{CT}\left[Ch+C\sum_{j<k}\left(h^2+\frac h{j+1}\right)\right]
 \leq C\eta_h.
\]
Returning to $v=u-hg(x)$ preserves this bound in $\|\cdot\|_*$, proving the state and initialization-derivative estimate.

\paragraph{Time differences.}
The discrete equations give the exact identities
\[
 \frac{x_{k+1}-x_k}h=v_{k+1},\qquad
 \frac{v_{k+1}-v_k}h=-\frac3{t_k+2h}v_k-g(y_k).
\]
For $t_k\geq\delta$, the state estimate and the continuous bounds imply
$\|v_{k+1}-v(t_k)\|+\|v_k-v(t_k)\|+\|y_k-x(t_k)\|\leq C(\eta_h+h)$.
The coefficients $3/(t_k+2h)$ and $3/(t_k+\tau)$ differ by $O_\delta(h)$, and the remaining $h$-weighted terms in \eqref{eq:app-nagc-ode} are $O_\delta(h)$. Comparing the two equations proves \eqref{eq:app-nagc-time-derivative-consistency}.
\end{proof}

\paragraph{Scores.}
All score bounds below are uniform for policy parameters in a fixed bounded set. Assume the $C^2$ score $S$ has globally bounded first and second derivatives in all its arguments for $t\in[0,T]$, and set
$V(t;a,\theta)=S(t,Z(t;a),a;\theta)$ and
$\bar V_k(a,\theta)=S(t_k,\bar Z_k(a),a;\theta)$.
Writing $\xi=(a,\theta)$, the chain rule gives
\[
 |\bar V_k-V(t_k)|+\|D_\xi\bar V_k-D_\xi V(t_k)\|\leq C\eta_h.
\]
The mean-value formula applied to $S$ along the segment from $(t_k,\bar Z_k)$ to $(t_{k+1},\bar Z_{k+1})$, together with \eqref{eq:app-nagc-time-derivative-consistency}, gives
$| (\bar V_{k+1}-\bar V_k)/h-\dot V(t_k)|\leq C_\delta\eta_h$ for $t_k\geq\delta$.

For an integrated score, assume the $C^2$ field $\cAv$ has globally bounded first and second derivatives in all its arguments for $t\in[0,T]$, and let
\[
 \dot V=-\cAv(Z,V,t;\theta),\qquad
 \bar V_{k+1}=\bar V_k-h\cAv(\bar Z_k,\bar V_k,t_k;\theta),\qquad
 V(0)=\bar V_0=1.
\]
Global Lipschitz continuity of $\cAv$ and Gronwall's inequality bound $V,\bar V_k$ and their first initialization/parameter derivatives on $[0,T]$. Since $\|\dot Z\|+\|D_a\dot Z\|\leq C$, the equations for $V$ and $D_\xi V$ have Euler residuals $O(h^2)$, uniformly down to time zero. For
$E_k^V=|\bar V_k-V(t_k)|+\|D_\xi\bar V_k-D_\xi V(t_k)\|$,
\[
 E_0^V=0,\qquad E_{k+1}^V\leq(1+Ch)E_k^V+Ch\eta_h+Ch^2,
 \qquad E_k^V\leq C\eta_h.
\]
The update then gives $| (\bar V_{k+1}-\bar V_k)/h-\dot V(t_k)|\leq C\eta_h$. Twice differentiating the score equation yields a linear equation for $D_\xi^2V$ with bounded forcing, so $\|D_\xi^2V\|\leq C$. For either score construction, first derivatives in $(t,a,\theta)$ are uniformly bounded and second derivatives are bounded by $C_\delta$ for $t\geq\delta$, as required by \cref{ass:disc_consistency-detailed}.

\paragraph{Application to restarted trajectories.}
Apply the proposition and use the phase-start position $a$ as the abstract state coordinate, with $\Fx(t;a)=x(t;a)$, $\Fbarx(k;a)=x_k(a)$, and $R$ the identity on positions. Each phase internally initializes its velocity with $hb_c(a)$ or $hb_d(a)$; its score may observe the full position--velocity state. Thus the initialization derivative is precisely $D_a$, and \cref{thm:disc_cont_close-detailed} applies with $\epsilon_h=O(h\log(1/h))$.
\endgroup

%% file: main.bbl
\begin{thebibliography}{46}
\providecommand{\natexlab}[1]{#1}
\providecommand{\url}[1]{\texttt{#1}}
\expandafter\ifx\csname urlstyle\endcsname\relax
  \providecommand{\doi}[1]{doi: #1}\else
  \providecommand{\doi}{doi: \begingroup \urlstyle{rm}\Url}\fi

\bibitem[Andrychowicz et~al.(2016)Andrychowicz, Denil, Colmenarejo, Hoffman,
  Pfau, Schaul, Shillingford, and de~Freitas]{l2o-dm}
Marcin Andrychowicz, Misha Denil, Sergio~Gomez Colmenarejo, Matthew~W. Hoffman,
  David Pfau, Tom Schaul, Brendan Shillingford, and Nando de~Freitas.
\newblock Learning to learn by gradient descent by gradient descent.
\newblock In Daniel~D. Lee, Masashi Sugiyama, Ulrike von Luxburg, Isabelle
  Guyon, and Roman Garnett (eds.), \emph{Advances in Neural Information
  Processing Systems 29: Annual Conference on Neural Information Processing
  Systems 2016, December 5-10, 2016, Barcelona, Spain}, pp.\  3981--3989, 2016.
\newblock URL
  \url{https://proceedings.neurips.cc/paper/2016/hash/fb87582825f9d28a8d42c5e5e5e8b23d-Abstract.html}.

\bibitem[Applegate et~al.(2021)Applegate, D{\'\i}az, Hinder, Lu, Lubin,
  O'Donoghue, and Schudy]{applegate2021practical}
David Applegate, Mateo D{\'\i}az, Oliver Hinder, Haihao Lu, Miles Lubin,
  Brendan O'Donoghue, and Warren Schudy.
\newblock Practical large-scale linear programming using primal-dual hybrid
  gradient.
\newblock \emph{Advances in Neural Information Processing Systems},
  34:\penalty0 20243--20257, 2021.

\bibitem[Applegate et~al.(2023)Applegate, Hinder, Lu, and
  Lubin]{applegate2023faster}
David Applegate, Oliver Hinder, Haihao Lu, and Miles Lubin.
\newblock Faster first-order primal-dual methods for linear programming using
  restarts and sharpness.
\newblock \emph{Mathematical Programming}, 201\penalty0 (1):\penalty0 133--184,
  2023.

\bibitem[Applegate et~al.(2025)Applegate, D{\'\i}az, Hinder, Lu, Lubin,
  O'Donoghue, and Schudy]{applegate2025pdlp}
David Applegate, Mateo D{\'\i}az, Oliver Hinder, Haihao Lu, Miles Lubin,
  Brendan O'Donoghue, and Warren Schudy.
\newblock Pdlp: A practical first-order method for large-scale linear
  programming.
\newblock \emph{arXiv preprint arXiv:2501.07018}, 2025.

\bibitem[Arrow et~al.(1958)Arrow, Hurwicz, and Uzawa]{arrow1958studies}
Kenneth~Joseph Arrow, Leonid Hurwicz, and Hirofumi Uzawa.
\newblock \emph{Studies in linear and non-linear programming}, volume~2.
\newblock Stanford University Press Stanford, 1958.

\bibitem[Beck(2017)]{beck2017first}
Amir Beck.
\newblock \emph{First-order methods in optimization}.
\newblock SIAM, 2017.

\bibitem[Bengio et~al.(2013)Bengio, L{\'e}onard, and
  Courville]{bengio2013estimating}
Yoshua Bengio, Nicholas L{\'e}onard, and Aaron Courville.
\newblock Estimating or propagating gradients through stochastic neurons for
  conditional computation.
\newblock \emph{arXiv preprint arXiv:1308.3432}, 2013.

\bibitem[Bertrand et~al.(2022)Bertrand, Klopfenstein, Massias, Blondel, Vaiter,
  Gramfort, and Salmon]{bertrand2022implicit}
Quentin Bertrand, Quentin Klopfenstein, Mathurin Massias, Mathieu Blondel,
  Samuel Vaiter, Alexandre Gramfort, and Joseph Salmon.
\newblock Implicit differentiation for fast hyperparameter selection in
  non-smooth convex learning.
\newblock \emph{Journal of Machine Learning Research}, 23\penalty0
  (149):\penalty0 1--43, 2022.

\bibitem[Blondel et~al.(2022)Blondel, Berthet, Cuturi, Frostig, Hoyer,
  Llinares-L{\'o}pez, Pedregosa, and Vert]{blondel2022efficient}
Mathieu Blondel, Quentin Berthet, Marco Cuturi, Roy Frostig, Stephan Hoyer,
  Felipe Llinares-L{\'o}pez, Fabian Pedregosa, and Jean-Philippe Vert.
\newblock Efficient and modular implicit differentiation.
\newblock In \emph{Advances in Neural Information Processing Systems},
  volume~35, pp.\  5230--5242, 2022.

\bibitem[Bolte et~al.(2022)Bolte, Le, Pauwels, and
  Silveti-Falls]{bolte2022nonsmoothimplicitdifferentiationmachine}
Jérôme Bolte, Tam Le, Edouard Pauwels, and Antonio Silveti-Falls.
\newblock Nonsmooth implicit differentiation for machine learning and
  optimization, 2022.
\newblock URL \url{https://arxiv.org/abs/2106.04350}.

\bibitem[Bubba et~al.(2017)Bubba, Juvonen, Lehtonen, M{\"a}rz, Meaney, Purisha,
  and Siltanen]{cheesedata}
Tatiana~A. Bubba, Markus Juvonen, Jonatan Lehtonen, Maximilian M{\"a}rz,
  Alexander Meaney, Zenith Purisha, and Samuli Siltanen.
\newblock Tomographic {X}-ray data of carved cheese, 2017.
\newblock URL \url{https://doi.org/10.5281/zenodo.1254210}.

\bibitem[Chambolle \& Pock(2011)Chambolle and Pock]{chambolle2011first}
Antonin Chambolle and Thomas Pock.
\newblock A first-order primal-dual algorithm for convex problems with
  applications to imaging.
\newblock \emph{Journal of mathematical imaging and vision}, 40\penalty0
  (1):\penalty0 120--145, 2011.

\bibitem[Chang \& Lin(2011)Chang and Lin]{libsvm}
Chih-Chung Chang and Chih-Jen Lin.
\newblock {LIBSVM}: A library for support vector machines.
\newblock \emph{ACM Transactions on Intelligent Systems and Technology},
  2:\penalty0 27:1--27:27, 2011.
\newblock Software available at \url{http://www.csie.ntu.edu.tw/~cjlin/libsvm}.

\bibitem[Chen et~al.(2020)Chen, Amos, and Nickel]{chen2020learning}
Ricky~TQ Chen, Brandon Amos, and Maximilian Nickel.
\newblock Learning neural event functions for ordinary differential equations.
\newblock \emph{arXiv preprint arXiv:2011.03902}, 2020.

\bibitem[Chen et~al.(2022)Chen, Chen, Chen, Heaton, Liu, Wang, and
  Yin]{chen2022learning}
Tianlong Chen, Xiaohan Chen, Wuyang Chen, Howard Heaton, Jialin Liu, Zhangyang
  Wang, and Wotao Yin.
\newblock Learning to optimize: A primer and a benchmark.
\newblock \emph{Journal of Machine Learning Research}, 23\penalty0
  (189):\penalty0 1--59, 2022.

\bibitem[Chen et~al.(2024)Chen, Liu, and Yin]{chen2024learning}
Xiaohan Chen, Jialin Liu, and Wotao Yin.
\newblock Learning to optimize: A tutorial for continuous and mixed-integer
  optimization.
\newblock \emph{Science China Mathematics}, 67\penalty0 (6):\penalty0
  1191--1262, 2024.

\bibitem[Cherukuri et~al.(2016)Cherukuri, Mallada, and
  Cort{\'e}s]{cherukuri2016asymptotic}
Ashish Cherukuri, Enrique Mallada, and Jorge Cort{\'e}s.
\newblock Asymptotic convergence of constrained primal--dual dynamics.
\newblock \emph{Systems \& Control Letters}, 87:\penalty0 10--15, 2016.

\bibitem[Feijer \& Paganini(2010)Feijer and Paganini]{feijer2010stability}
Diego Feijer and Fernando Paganini.
\newblock Stability of primal--dual gradient dynamics and applications to
  network optimization.
\newblock \emph{Automatica}, 46\penalty0 (12):\penalty0 1974--1981, 2010.

\bibitem[Fercoq(2022)]{fercoq2022quadratic}
Olivier Fercoq.
\newblock Quadratic error bound of the smoothed gap and the restarted averaged
  primal-dual hybrid gradient.
\newblock \emph{arXiv preprint arXiv:2206.03041}, 2022.

\bibitem[Fercoq \& Qu(2019)Fercoq and Qu]{fercoq2019adaptive}
Olivier Fercoq and Zheng Qu.
\newblock Adaptive restart of accelerated gradient methods under local
  quadratic growth condition.
\newblock \emph{IMA Journal of Numerical Analysis}, 39\penalty0 (4):\penalty0
  2069--2095, 2019.

\bibitem[Feurer \& Hutter(2019)Feurer and Hutter]{feurer2019hyperparameter}
Matthias Feurer and Frank Hutter.
\newblock Hyperparameter optimization.
\newblock In \emph{Automated Machine Learning}, pp.\  3--33. Springer
  International Publishing, 2019.

\bibitem[Giselsson \& Boyd(2014)Giselsson and Boyd]{giselsson2014monotonicity}
Pontus Giselsson and Stephen Boyd.
\newblock Monotonicity and restart in fast gradient methods.
\newblock In \emph{53rd IEEE Conference on Decision and Control}, pp.\
  5058--5063. IEEE, 2014.

\bibitem[Goebel et~al.(2009)Goebel, Sanfelice, and Teel]{goebel2009hybrid}
Rafal Goebel, Ricardo~G Sanfelice, and Andrew~R Teel.
\newblock Hybrid dynamical systems.
\newblock \emph{IEEE control systems magazine}, 29\penalty0 (2):\penalty0
  28--93, 2009.

\bibitem[H{\"a}m{\"a}l{\"a}inen et~al.(2015)H{\"a}m{\"a}l{\"a}inen, Harhanen,
  Kallonen, Kujanp{\"a}{\"a}, Niemi, and Siltanen]{walnutdata}
Keijo H{\"a}m{\"a}l{\"a}inen, Lauri Harhanen, Aki Kallonen, Antti
  Kujanp{\"a}{\"a}, Esa Niemi, and Samuli Siltanen.
\newblock Tomographic {X}-ray data of a walnut, 2015.
\newblock URL \url{https://doi.org/10.5281/zenodo.1254206}.

\bibitem[Helmke \& Moore(2012)Helmke and Moore]{helmke2012optimization}
Uwe Helmke and John~B Moore.
\newblock \emph{Optimization and dynamical systems}.
\newblock Springer Science \& Business Media, 2012.

\bibitem[Hinder(2024)]{hinder2024worst}
Oliver Hinder.
\newblock Worst-case analysis of restarted primal-dual hybrid gradient on
  totally unimodular linear programs.
\newblock \emph{Operations Research Letters}, 57:\penalty0 107199, 2024.

\bibitem[Laborde \& Oberman(2020)Laborde and Oberman]{laborde2020lyapunov}
Maxime Laborde and Adam Oberman.
\newblock A lyapunov analysis for accelerated gradient methods: From
  deterministic to stochastic case.
\newblock In \emph{International conference on artificial intelligence and
  statistics}, pp.\  602--612. PMLR, 2020.

\bibitem[Li \& Shi(2024)Li and Shi]{li2024understanding}
Bowen Li and Bin Shi.
\newblock Understanding the pdhg algorithm via high-resolution differential
  equations.
\newblock \emph{arXiv preprint arXiv:2403.11139}, 2024.

\bibitem[Liu et~al.(2023)Liu, Chen, Wang, Yin, and Cai]{liu2023towards}
Jialin Liu, Xiaohan Chen, Zhangyang Wang, Wotao Yin, and HanQin Cai.
\newblock Towards constituting mathematical structures for learning to
  optimize.
\newblock In \emph{Proceedings of the 40th International Conference on Machine
  Learning}, pp.\  21426--21449, 2023.

\bibitem[Lu \& Yang(2025)Lu and Yang]{lu2025geometry}
Haihao Lu and Jinwen Yang.
\newblock On the geometry and refined rate of primal--dual hybrid gradient for
  linear programming.
\newblock \emph{Mathematical Programming}, 212\penalty0 (1):\penalty0 349--387,
  2025.

\bibitem[Lv et~al.(2017)Lv, Jiang, and Li]{rnnprop}
Kaifeng Lv, Shunhua Jiang, and Jian Li.
\newblock Learning gradient descent: Better generalization and longer horizons.
\newblock In Doina Precup and Yee~Whye Teh (eds.), \emph{Proceedings of the
  34th International Conference on Machine Learning, {ICML} 2017, Sydney, NSW,
  Australia, 6-11 August 2017}, volume~70 of \emph{Proceedings of Machine
  Learning Research}, pp.\  2247--2255. {PMLR}, 2017.
\newblock URL \url{http://proceedings.mlr.press/v70/lv17a.html}.

\bibitem[Moucer et~al.(2023)Moucer, Taylor, and Bach]{moucer2023systematic}
C{\'e}line Moucer, Adrien Taylor, and Francis Bach.
\newblock A systematic approach to lyapunov analyses of continuous-time models
  in convex optimization.
\newblock \emph{SIAM Journal on Optimization}, 33\penalty0 (3):\penalty0
  1558--1586, 2023.

\bibitem[O'Donoghue \& Cand{\`e}s(2015)O'Donoghue and
  Cand{\`e}s]{o2015adaptive}
Brendan O'Donoghue and Emmanuel Cand{\`e}s.
\newblock Adaptive restart for accelerated gradient schemes.
\newblock \emph{Foundations of Computational Mathematics}, 15\penalty0
  (3):\penalty0 715--732, Jun 2015.
\newblock ISSN 1615-3383.
\newblock \doi{10.1007/s10208-013-9150-3}.
\newblock URL \url{https://doi.org/10.1007/s10208-013-9150-3}.

\bibitem[Pilbeam(2018)]{pilbeam2018finance}
Keith Pilbeam.
\newblock \emph{Finance and financial markets}.
\newblock Bloomsbury Publishing, 2018.

\bibitem[Rahimi \& Recht(2007)Rahimi and Recht]{rahimi2007random}
Ali Rahimi and Benjamin Recht.
\newblock Random features for large-scale kernel machines.
\newblock In \emph{Advances in Neural Information Processing Systems},
  volume~20, 2007.

\bibitem[Shi et~al.(2022)Shi, Du, Jordan, and Su]{shi2022understanding}
Bin Shi, Simon~S Du, Michael~I Jordan, and Weijie~J Su.
\newblock Understanding the acceleration phenomenon via high-resolution
  differential equations.
\newblock \emph{Mathematical Programming}, 195:\penalty0 79--148, 2022.

\bibitem[Su et~al.(2016)Su, Boyd, and Cand{{\`e}}s]{su2016differential}
Weijie Su, Stephen Boyd, and Emmanuel~J. Cand{{\`e}}s.
\newblock A differential equation for modeling nesterov's accelerated gradient
  method: Theory and insights.
\newblock \emph{Journal of Machine Learning Research}, 17\penalty0
  (153):\penalty0 1--43, 2016.
\newblock URL \url{http://jmlr.org/papers/v17/15-084.html}.

\bibitem[Taha(1997)]{taha1997operations}
Hamdy~A Taha.
\newblock \emph{Operations research: an introduction}.
\newblock Prentice hall Upper Saddle River, NJ, 6 edition, 1997.

\bibitem[Van Der~Schaft \& Schumacher(2007)Van Der~Schaft and
  Schumacher]{van2007introduction}
Arjan~J Van Der~Schaft and Hans Schumacher.
\newblock \emph{An introduction to hybrid dynamical systems}, volume 251.
\newblock springer, 2007.

\bibitem[Wibisono et~al.(2016)Wibisono, Wilson, and
  Jordan]{wibisono2016variational}
Andre Wibisono, Ashia~C Wilson, and Michael~I Jordan.
\newblock A variational perspective on accelerated methods in optimization.
\newblock \emph{proceedings of the National Academy of Sciences}, 113\penalty0
  (47):\penalty0 E7351--E7358, 2016.

\bibitem[Xie et~al.(2024)Xie, Yin, and Wen]{xie2024ode}
Zhonglin Xie, Wotao Yin, and Zaiwen Wen.
\newblock {ODE-based Learning to Optimize}, 2024.
\newblock URL \url{https://arxiv.org/abs/2406.02006}.

\bibitem[Xie et~al.(2025)Xie, Fong, Yuan, and Wen]{xie2025accelerating}
Zhonglin Xie, Yiman Fong, Haoran Yuan, and Zaiwen Wen.
\newblock Accelerating optimization via differentiable stopping time.
\newblock \emph{arXiv preprint arXiv:2505.22509}, 2025.

\bibitem[Xu et~al.(2023)Xu, Molloy, and Gould]{xu2023revisiting}
Ming Xu, Timothy~L Molloy, and Stephen Gould.
\newblock Revisiting implicit differentiation for learning problems in optimal
  control.
\newblock In \emph{Advances in Neural Information Processing Systems},
  volume~36, pp.\  60060--60076, 2023.

\bibitem[Yang et~al.(2023{\natexlab{a}})Yang, Chen, Zhu, He, Tao, Liang, and
  Wang]{yang2023learninggeneralizeprovablylearning}
Junjie Yang, Tianlong Chen, Mingkang Zhu, Fengxiang He, Dacheng Tao, Yingbin
  Liang, and Zhangyang Wang.
\newblock Learning to generalize provably in learning to optimize,
  2023{\natexlab{a}}.
\newblock URL \url{https://arxiv.org/abs/2302.11085}.

\bibitem[Yang et~al.(2023{\natexlab{b}})Yang, Chen, Chen, Wang, and
  Liang]{yang2023ml2o}
Junjie Yang, Xuxi Chen, Tianlong Chen, Zhangyang Wang, and Yingbin Liang.
\newblock M-l2o: Towards generalizable learning-to-optimize by test-time fast
  self-adaptation.
\newblock \emph{arXiv preprint arXiv:2303.00039}, 2023{\natexlab{b}}.

\bibitem[Zhao et~al.(2025)Zhao, Zhou, Li, Tang, Wang, Hou, Min, Zhang, Zhang,
  Dong, Du, Yang, Chen, Chen, Jiang, Ren, Li, Tang, Liu, Liu, Nie, and
  Wen]{zhao2025surveylargelanguagemodels}
Wayne~Xin Zhao, Kun Zhou, Junyi Li, Tianyi Tang, Xiaolei Wang, Yupeng Hou,
  Yingqian Min, Beichen Zhang, Junjie Zhang, Zican Dong, Yifan Du, Chen Yang,
  Yushuo Chen, Zhipeng Chen, Jinhao Jiang, Ruiyang Ren, Yifan Li, Xinyu Tang,
  Zikang Liu, Peiyu Liu, Jian-Yun Nie, and Ji-Rong Wen.
\newblock A survey of large language models, 2025.
\newblock URL \url{https://arxiv.org/abs/2303.18223}.

\end{thebibliography}
